\documentclass{article} 
\usepackage{iclr2027_conference,times}
\iclrfinalcopy 

\usepackage{amsmath,amsfonts,bm}

\def\eqref#1{equation~\ref{#1}}

\def\1{\bm{1}}

\DeclareMathAlphabet{\mathsfit}{\encodingdefault}{\sfdefault}{m}{sl}
\SetMathAlphabet{\mathsfit}{bold}{\encodingdefault}{\sfdefault}{bx}{n}

\newcommand{\KL}{D_{\mathrm{KL}}}

\usepackage{hyperref}
\usepackage{url}

\usepackage{amsmath,amssymb,amsthm}
\usepackage{dsfont}
\usepackage{booktabs}
\usepackage{algorithm}
\usepackage{algpseudocode}
\usepackage{graphicx}
\usepackage{microtype}
\usepackage{multirow}
\usepackage{url}
\usepackage{hyperref}
\usepackage{thm-restate}
\usepackage{xcolor}
\usepackage{graphicx}
\usepackage{arydshln}
\usepackage{graphicx,subcaption}
\usepackage{xspace}
\usepackage{wrapfig}
\hypersetup{hidelinks}
\DeclareMathSizes{9}{10}{7}{5}
\DeclareMathSizes{8}{10}{7}{5}

\newtheorem{theorem}{Theorem}
\newtheorem{proposition}{Proposition}

\newtheorem{lemma}{Lemma}

\newcommand{\HV}{\operatorname{HV}}

\newcommand{\ind}{\mathds{1}}

\newcommand{\rref}{\mathbf r^{\mathrm{ref}}}

\newcommand{\rvec}{\mathbf r}

\newcommand{\diagnosticfigure}[1]{%
  \IfFileExists{#1}{\includegraphics[width=\linewidth]{#1}}{%
    \fbox{\parbox[c][3cm][c]{0.9\linewidth}{\centering
      \textcolor{magenta}{Figure pending}\\[0.5em]
      \texttt{\detokenize{#1}}}}%
  }%
}

\newcommand{\AlgNameShort}{\textsc{\small{HVFT}}\xspace}
\newcommand{\AlgNameDef}{Hypervolume Fine-Tuning (\textsc{\small{HVFT}})\xspace}
\newcommand{\AlgNameNFT}{\textsc{\small{DiffusionNFT}}\xspace}
\newcommand{\AlgNameBlend}{\textsc{\small{DiffusionBlend}}\xspace}
\newcommand{\AlgNamePep}{\textsc{\small{PepTune}}\xspace}
\newcommand{\AlgNameTR}{\textsc{\small{TR2D2}}\xspace}

\usepackage{cleveref}
\crefname{equation}{Equation}{Equations}
\Crefname{equation}{Equation}{Equations}
\crefname{appendix}{Appendix}{Appendices}
\Crefname{appendix}{Appendix}{Appendices}

\definecolor{pastelblueold}{RGB}{56,146,236}
\definecolor{pastelblue}{RGB}{43,115,187}
\definecolor{pastelgreen}{RGB}{63,159,95}

\definecolor{oxfordblue}{HTML}{163A70}
\definecolor{softbluebg}{HTML}{EAF3FF}
\definecolor{deepviolet}{HTML}{4B2E83}
\definecolor{coolgraybg}{HTML}{F3F4F7}

\hypersetup{
  colorlinks=true,
  citecolor=pastelgreen,
  linkcolor=pastelblue,    
  urlcolor=pastelblue      
}

\title{Multi-Objective Generative Optimization via\\ Expected Hypervolume Fine-Tuning}

\author{Juan Guevara$^{1}$, Riccardo De Santi$^{2,3}$, Andreas Krause$^{2,3}$, Raul Astudillo$^{1}$\\
\normalfont $^{1}$Mohamed Bin Zayed University of Artificial Intelligence \quad $^{2}$ETH Zurich \quad $^{3}$ETH AI Center}

\begin{document}
\maketitle
\lhead{Preprint}


\vspace{-0.2cm}

\begin{abstract}
    Scientific discovery requires plausible designs spanning useful trade-offs among competing properties: a molecule with strong binding affinity, for example, may be difficult to synthesize.
    Pretrained generative models, e.g., flow and diffusion generators, provide a foundation for proposing such designs, but steering them toward high-quality, complementary trade-offs remains challenging.
    Inference-time search incurs high per-sample costs, while fine-tuning typically relies on fixed per-candidate rewards that do not capture how designs complement one another.
    We introduce \AlgNameDef, which adapts a single pretrained generator to maximize the expected hypervolume of its samples while controlling departure from the pretrained distribution.
    Expected hypervolume jointly rewards candidate quality and collective trade-off coverage.
    Nonetheless, we show that it cannot be expressed as the expectation of a fixed per-candidate reward and therefore cannot be optimized directly with standard techniques.
    A key observation of this work is that its first variation corresponds to the expected marginal hypervolume contribution of a candidate, which yields a tractable adaptive reward via an unbiased estimator based on model samples.
    \AlgNameShort uses this reward to decompose the objective into a sequence of
    KL-proximal scalar-reward subproblems.
    We establish concavity, relative smoothness, and convergence guarantees for the resulting distributional procedure.
    Across therapeutic-peptide and molecular design tasks spanning discrete diffusion and continuous flow-matching models, \AlgNameShort achieves state-of-the-art expected hypervolume and trade-off coverage.
\end{abstract}

\section{Introduction}

Scientific discovery often requires balancing competing objectives. A
peptide with high binding affinity to a protein, for example,
may have low cell permeability, and vice versa
\citep{tang2025peptune}. The most useful compromise may remain unknown
until experimental evidence or downstream preferences become available.
A useful generator should therefore produce candidate sets that
collectively represent high-quality, complementary trade-offs.

\looseness -1 Pretrained generative models provide a prior over plausible designs,
but their sampling distributions are not optimized for the properties
of interest. Generative optimization adapts these distributions toward
improved objective values
\citep{domingoenrich2024adjoint,desanti2025fdc, gutjahr2025constrained, kharitenko2025landing}. In the multi-objective
setting, however, improving candidates individually is not enough:
samples must also complement one another across competing objectives. The key
challenge is therefore to optimize this collective behavior, rather
than a fixed score assigned to each candidate in isolation.

\begin{figure}[!t]
    \centering
    \begin{minipage}{\linewidth}
        \centering
        \begin{subfigure}[t]{0.646\linewidth}
            \centering
            \includegraphics[width=\linewidth]{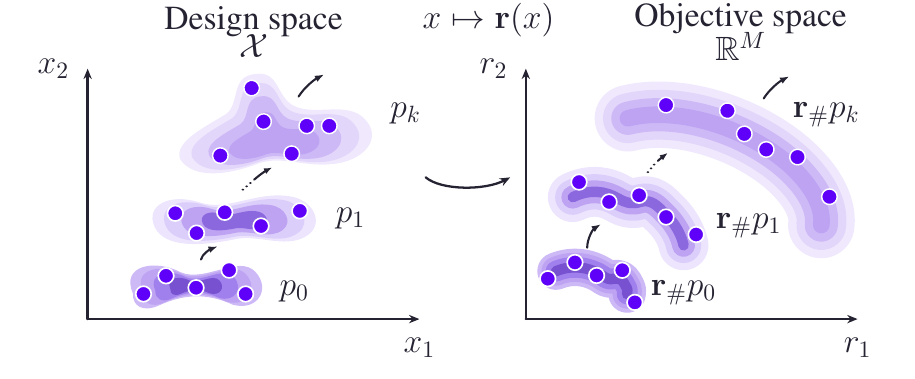}
            \caption{Sequential fine-tuning.}
            \label{fig:schematic-ft}
        \end{subfigure}\hfill
        \begin{subfigure}[t]{0.323\linewidth}
            \centering
            \includegraphics[width=\linewidth]{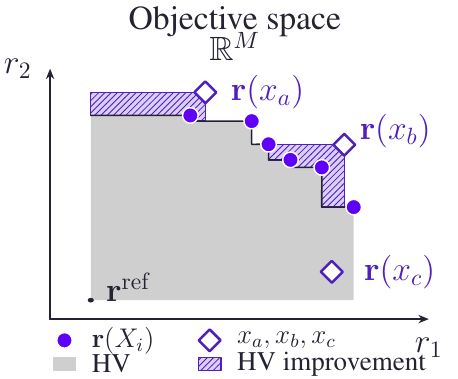}
            \caption{Marginal hypervolume reward.}
            \label{fig:hypervolume-geometry}
        \end{subfigure}
    \end{minipage}
    \caption{\textbf{\AlgNameShort for multi-objective generative fine-tuning.}
(a) \AlgNameShort iteratively constructs a scalar reward from the current
distribution $p_k$ and fine-tunes the generator, shifting the pushforward
distribution $\mathbf r_{\#}p_k$ toward high-quality, complementary
trade-offs.
(b) The gray region is the hypervolume dominated relative to
$\mathbf r^{\mathrm{ref}}$. Hatching shows the marginal contributions of
$x_a$ and $x_b$, while the dominated candidate $x_c$ contributes zero.
\AlgNameShort estimates these contributions against background sets from $p_k$ to
define the next fine-tuning reward.}
    \label{fig:hvft-overview}
\end{figure}

Existing methods provide several forms of multi-objective steering.
Inference-time search and guidance can seek favorable trade-offs, but
they require additional optimization whenever candidates are generated
\citep{tang2025peptune,yao2024proud,annadani2025preference} and, without
updating the model, may have limited ability to depart from the
pretrained prior \citep{Klarner2024ContextGuidedDF}. Fine-tuning amortizes this cost, yet it faces other limitations. Fixed scalarizations
score candidates independently
\citep{tang2025tr2d2,mahood2026moose,patel2026fullLength,Lin2026FineTuningDM};
preference-conditioned and composed models require specifying a preference distribution
\citep{golan2026paretoslider,Cheng2025DiffusionBI}; and dominance-based
signals neither distinguish mutually nondominated candidates nor
measure how each contributes to the generated set
\citep{drugflow2025}. No algorithm directly trains the sampling distribution for
both candidate quality and collective trade-off coverage.



Motivated by this challenge, we address the following question:

\begin{center}
\emph{How can we fine-tune a generative model to directly produce
samples that jointly achieve high objective quality and broad trade-off
coverage?}
\end{center}

We introduce \AlgNameDef, which fine-tunes a
single generative model to maximize the expected hypervolume of its
generated candidate sets while controlling departure from the pretrained
distribution. Hypervolume measures the objective-space region dominated
by a candidate set relative to a fixed reference point
\citep{zitzler1999multiobjective,guerreiro2021hypervolume}. It therefore
rewards both strong individual candidates and complementary trade-off
coverage, without requiring a distribution over preferences.

\looseness -1 Unlike standard fine-tuning objectives, expected hypervolume cannot be
expressed as the expectation of a fixed per-candidate reward. A
candidate's value depends on the objective-space regions covered by the
other samples, so its reward must adapt to the current sampling
distribution
\citep{domingoenrich2024adjoint,Zheng2025DiffusionNFTOD,Zhu2025EnhancingRF,jensen2025value}.
\AlgNameShort addresses this limitation using Flow Density Control
\citep{desanti2025fdc}, which optimizes a distributional objective
through a sequence of KL-proximal subproblems driven by its first
variation (see Figure~\ref{fig:hvft-overview}). A key observation is that, for expected hypervolume, this
first variation is the expected marginal hypervolume contribution of a
candidate relative to samples from the current model. This identity
gives the reward a clear interpretation and makes it tractable to
estimate directly from model samples. At each outer iteration, \AlgNameShort
estimates this adaptive reward under the current distribution, uses it
with a model-specific scalar-reward solver, and recomputes it after the
update. We further establish concavity and relative smoothness of the
objective, yielding monotone improvement and convergence guarantees for
this procedure.

Across therapeutic-peptide
\citep{tang2025peptune,tang2025tr2d2} and molecular design
\citep{drugflow2025} tasks spanning discrete diffusion and continuous
flow-matching models, \AlgNameShort achieves state-of-the-art expected
hypervolume and trade-off coverage without additional inference-time
search.

Our main contributions are as follows:
\begin{itemize}
    \item We introduce expected hypervolume, a distributional training
criterion for multi-objective generative optimization that jointly rewards
quality and collective trade-off coverage.

\item We express the first variation of expected hypervolume in terms of
marginal hypervolume contributions and derive an unbiased sample-based
estimator. This yields \AlgNameShort, a sequence of scalar-reward fine-tuning
subproblems for which we establish convergence guarantees.

    \item We demonstrate that HVFT achieves state-of-the-art performance on
    therapeutic-peptide and molecular design tasks spanning discrete diffusion and
    continuous flow-matching models, improving the expected hypervolume
    of sampled sets without inference-time search.
\end{itemize}

\section{Related work}\label{sec:related}

We first discuss existing methods for multi-objective steering of
pretrained generators and then review the
methodological foundations of \AlgNameShort. We
conclude by briefly discussing adjacent multi-objective optimization
settings that target different problems.

\paragraph{Multi-objective steering of pretrained generators.}
Methods for steering pretrained generators toward multiple objectives
differ primarily in the information carried by their steering signal.

\textit{Ordinal Pareto signals.} Pareto dominance guides
inference-time search in PepTune \citep{tang2025peptune}, replay-buffer
construction and off-policy fine-tuning in TR2D2 and its extensions
\citep{tang2025tr2d2,mahood2026moose,patel2026fullLength}, and the
selection of preference pairs for direct preference optimization
\citep{drugflow2025}. Dominance records which candidates are preferable
but not by how much: it is invariant to strictly increasing
transformations of each objective, cannot order mutually nondominated
candidates, and does not quantify their contributions to a candidate
set. 

\textit{Scalarized per-candidate rewards.} A second family maps each
objective vector to a scalar utility and applies standard reward-based
fine-tuning, sometimes incorporating thresholds or predictive
uncertainty \citep{Lin2026FineTuningDM,Chen2025UncertaintyAwareMR}.
TR2-D2-based methods, for example, sum the objectives of candidates in
their nondominated replay buffer
\citep{tang2025tr2d2,mahood2026moose,patel2026fullLength}. Because a
fixed scalarization scores candidates independently and generally
favors a particular trade-off, repeated samples need not cover the
front.

\textit{Preference-conditioned generation.} A third family makes
the trade-off an input by conditioning on a preference vector during
training \citep{golan2026paretoslider} or inference
\citep{Chen2025MultiObjectiveGuidedDF}, or by composing separately
aligned models at sampling time \citep{Cheng2025DiffusionBI}. These
methods provide precise control when the desired trade-off is known.
Producing a set of alternatives instead requires choosing a distribution
over preferences, such as a grid or a Das--Dennis construction. This
choice becomes harder as the number of objectives grows and does not
directly optimize the coverage of the resulting samples. 

\textit{Positioning of \AlgNameShort.}
The preceding approaches rely on ordinal comparisons, fixed per-candidate
utilities, or externally specified preferences. \AlgNameShort instead introduces a
cardinal set-level training signal: each candidate is valued by the
hypervolume it contributes relative to samples from the current model. This
signal distinguishes mutually nondominated candidates, accounts for the
trade-offs already covered by the generator, and requires no specified
preference distribution.

\paragraph{Methodological foundations.}
We discuss generators and hypervolume as a set-level objective. We further describe the distributional optimization framework
that connects them.

\textit{Reward-based generative fine-tuning.}
Pretrained diffusion, flow-matching, and discrete-diffusion models
provide priors over plausible molecular structures and biological
sequences
\citep{ho2020denoising,lipman2023flowmatching,lou2024discrete,tang2025peptune,flomolDunn}.
Reward-based methods adapt these models toward target properties
\citep{domingoenrich2024adjoint,Liu2025FlowGRPOTF,Zheng2025DiffusionNFTOD}.
\AlgNameShort likewise adapts a pretrained model using a compatible reward-based
method as the inner solver within a sequential fine-tuning procedure,
with the marginal hypervolume reward recomputed as the sampling
distribution changes.

\textit{Hypervolume as an optimization signal.}
Hypervolume measures the region of objective space dominated by a
candidate set relative to a reference point
\citep{zitzler1999multiobjective,guerreiro2021hypervolume}; its
diminishing-returns structure relates it to submodular coverage
\citep{nemhauser1978analysis,krause2014submodular}. Beyond evaluation,
hypervolume has been used in Bayesian optimization, evolutionary selection,
Pareto-set learning, and reinforcement-learning
\citep{daulton2020differentiable,beume2007smsemoa,bader2011hype,zhang2023hypervolume,lu2025dynamic,steinberg2025amortized}.
These methods optimize the hypervolume of a finite set selected or
maintained by the algorithm. \AlgNameShort instead uses the expected hypervolume
of samples from the generator as its steering signal.

\textit{Distributional optimization of non-additive objectives.}
Set-level objectives make each candidate's value depend on the others.
Related work in submodular reinforcement learning (RL) captures these interactions through marginal-gain
estimators and successive additive approximations
\citep{prajapat2024submodular,desanti2024global}, while work in general utilities, or convex RL optimizes distributional
functionals via iterative linearizations
\citep{hazan2019provably,zahavy2021reward,mutti2023convex}. Flow Density Control provides
a general framework for optimizing nonlinear functionals of generative
distributions via generative model adaptation: at each outer iteration, it evaluates the first variation
at the current sampling distribution and uses it as a scalar reward in a
KL-proximal fine-tuning subproblem \citep{desanti2025fdc}. While prior work employed distributional flow fine-tuning to improve design-space coverage~\citep[e.g.,][]{de2025provable,de2026verifier,de2026unified}, here we tackle the problem of multi-objective optimization, i.e., a specific form of objective-space coverage.  \AlgNameShort uses this
framework to optimize expected hypervolume through sequential
scalar-reward fine-tuning. To make this possible, we derive the marginal
hypervolume first variation, construct an unbiased sample-based
estimator, and establish a relative-smoothness bound that controls the
step size.

\paragraph{Adjacent multi-objective optimization settings.}
Several related formulations differ in their evaluation access, prior information, and outputs.

\emph{Multi-objective Bayesian optimization} methods treat objective
evaluations as scarce and repeatedly select batches that improve the
hypervolume of an observed archive
\citep{daulton2020differentiable,stanton2022lambo,Gruver2023lambo2,jain2023multiobjective,steinberg2025amortized}.
\AlgNameShort instead assumes objectives can be queried at training scale and
returns a sampler that can subsequently generate candidates without
re-solving the optimization problem. \emph{Pareto-set approximation}
methods return an explicit approximation of the front or a map from
preferences to solutions, using, for example, hypervolume-based survival
selection \citep{beume2007smsemoa,bader2011hype} or learned Pareto-set
representations \citep{lin2022paretoset,zhang2023hypervolume}. They do
not use a pretrained distribution as a prior over plausible designs.
\emph{Offline multi-objective optimization} methods work from a fixed
dataset without further objective queries and steer generation using
learned predictors or dominance classifiers
\citep{yuan2025paretoflow,annadani2025preference}. \AlgNameShort instead queries
the objectives during training and uses only their evaluated values.

\section{Hypervolume Fine-Tuning}
\label{sec:method}

\looseness -1 Let $\mathcal X$ denote a design space and let
$\rvec(x)=(r_1(x),\ldots,r_M(x)) \in \mathbb{R}^M$ denote a vector of competing
objectives, all taken to be maximized. We assume that these objectives
can be evaluated on generated candidates, but they may be black-box and
nondifferentiable. Let $p_0$ denote a pretrained generative
distribution over $\mathcal X$. We seek to adapt this distribution so
that its samples collectively achieve high objective
quality and broad trade-off coverage while preserving useful behavior
learned during pretraining.

Achieving this goal requires a distributional objective that
values candidate sets and a way to optimize that objective via
generative model fine-tuning. The remainder of this section develops these
two ingredients. We first define expected hypervolume as a
distributional objective (\Cref{sec:hv-objective}) and then derive its
first variation, which is an expected marginal hypervolume contribution
(\Cref{sec:rewards}). This local reward leads to the sequential
fine-tuning procedure in \Cref{sec:sequential}.

\subsection{Expected hypervolume as a distributional objective}
\label{sec:hv-objective}

Evaluating samples collectively requires accounting for both objective
quality and complementary coverage. Hypervolume provides such a
measure: relative to a reference point $\rref\in\mathbb{R}^M$, it quantifies the
volume of objective space jointly dominated by a collection of
candidates
\citep{zitzler1999multiobjective,guerreiro2021hypervolume}.  Unlike a
fixed scalarization, hypervolume does not reduce the objectives to a
prescribed per-candidate score; unlike Pareto dominance alone, it
quantifies how much each candidate contributes to collective coverage.
The reference point and objective scaling determine the region and
relative volumes being measured, and we report these choices for each
task in Appendix~\ref{app:implementation}.

Writing $u\succeq v$ when $u_m\geq v_m$ for every objective $m$, we
define the hypervolume of a set $A\subseteq\mathcal X$ as
\begin{equation}
    \HV(A;\rref)
    :=
    \operatorname{Leb}_M
    \left(
        \bigcup_{x\in A}
        \left\{
            u\in\mathbb R^M:
            \rref\preceq u\preceq\rvec(x)
        \right\}
    \right),
    \label{eq:hv-definition}
\end{equation}
where $\operatorname{Leb}_M$ denotes $M$-dimensional Lebesgue measure
and $\HV(\varnothing;\rref)=0$. A candidate contributes only the
portion of its dominated region not already covered by the others.
Hypervolume therefore rewards stronger objective values, which expand
the dominated region, and complementary trade-offs, which extend that
region in previously uncovered directions. See Figure~\ref{fig:hypervolume-geometry} for an illustration of hypervolume.

We lift hypervolume to a distributional training objective by taking
its expectation over $n$ independent samples:
\footnote{Throughout the analysis, we assume that
$(r_m(x)-r_m^{\mathrm{ref}})_+\leq C$ for some finite $C>0$ and all
$x,m$. This bounds the hypervolume by $C^M$ and ensures that the
expectation is finite.}
\begin{equation}
    F_n(p)
    :=
    \mathbb E_{X_{1:n}\sim p^{\otimes n}}
    \left[
        \HV\!\left(\{X_1,\ldots,X_n\};\rref\right)
    \right].
    \label{eq:expected-hv}
\end{equation}
The parameter $n$ determines how many samples interact within the
fine-tuning objective rather than how many samples may subsequently be
generated. Empirically, we find that training with modest values of
$n$ improves hypervolume even when evaluating larger numbers of samples
(Section~\ref{sec:experiments}).

To limit departure from the pretrained model, we penalize deviations
from the pretrained distribution. Let $\mathcal P(\mathcal X)$ denote
the set of probability distributions on $\mathcal X$. Our optimization target is
\begin{equation}
    p^\star
    \in
    \arg\max_{p\in\mathcal P(\mathcal X)}
    \left\{
        F_n(p)
        -
        \alpha\KL(p\Vert p_0)
    \right\},
    \label{eq:distributional-problem}
\end{equation}
where $\alpha\geq0$ balances hypervolume performance against fidelity to the pretrained distribution.

The optimization in \cref{eq:distributional-problem} is not a standard
fixed-reward fine-tuning problem \citep{domingoenrich2024adjoint,Liu2025FlowGRPOTF}. When $n>1$, the value of a candidate
depends on the other $n-1$ samples and hence on the sampling
distribution itself. Thus, $F_n$ is generally nonlinear in $p$ and
cannot be written as the expectation of a fixed per-candidate reward.

\AlgNameShort handles this nonadditivity locally. At iteration $k$, we linearize
the objective around the current distribution $p_k$. Its first
variation assigns each candidate a scalar value equal to its first-order
effect on the objective. We optimize this local reward within a
KL-controlled neighborhood of $p_k$ to obtain $p_{k+1}$, recompute the
reward at the new distribution, and repeat. The next subsection derives
this reward; \Cref{sec:sequential} then uses it to define the sequence of
fine-tuning subproblems.

\subsection{First variation and the marginal hypervolume reward}
\label{sec:rewards}

We first derive the scalar reward that locally represents changes in
$F_n$. Following the distributional optimization framework of
\citet{desanti2025fdc}, a first variation of $F_n$ at $p$ is a function
$f_{n,p}$ such that, for every admissible distribution $q$,
\begin{equation}
F_n\!\left((1-\epsilon)p+\epsilon q\right)
=
F_n(p)
+
\epsilon\left(
\mathbb E_q[f_{n,p}(X)]
-
\mathbb E_p[f_{n,p}(X)]
\right)
+
o(\epsilon),
\qquad \epsilon\downarrow0.
\label{eq:first-variation}
\end{equation}
At the current iterate $p=p_k$, both $F_n(p_k)$ and
$\mathbb E_{p_k}[f_{n,p_k}(X)]$ are constant with respect to $q$.
Maximizing the linear approximation over $q$ therefore reduces to
maximizing $\mathbb E_q[f_{n,p_k}(X)]$, which has the additive form
required by a scalar-reward fine-tuning solver.

For expected hypervolume, this local reward has a direct
interpretation. Define the marginal hypervolume contribution of a
candidate $x\in\mathcal X$ relative to a background set
$A\subseteq\mathcal X$ as
\begin{equation}
    \Delta_{\HV}(x;A)
    :=
    \HV(A\cup\{x\};\rref)
    -
    \HV(A;\rref).
    \label{eq:marginal-hv}
\end{equation}

\begin{restatable}[Marginal hypervolume reward]{proposition}{marginalhvreward}
\label{thm:variation}
A first variation of $F_n$ at
$p$ is given by
\begin{equation}
    f_{n,p}(x)
    =
    n\,
    \mathbb E_{A\sim p^{\otimes(n-1)}}
    \left[
        \Delta_{\HV}(x;A)
    \right],
    \label{eq:variation-hvi}
\end{equation}
where $A$ is formed from $n-1$ independent draws from $p$.
\end{restatable}

\cref{eq:variation-hvi} rewards a candidate for the objective
space it covers beyond a background set drawn from the current model (see Figure~\ref{fig:hypervolume-geometry}).
The reward is large when the candidate has strong objective values and
adds a trade-off that current samples are unlikely to cover. It is not a
fixed scalarization of $\rvec(x)$: the background distribution changes
with $p$, so the reward changes as the model improves. Candidate
interactions are therefore retained in the outer loop even though the
inner solver receives a scalar reward. The proof of
Proposition~\ref{thm:variation} is given in Appendix~\ref{app:proofs}.

\subsection{Sequential fine-tuning via marginal hypervolume rewards}
\label{sec:sequential}

Flow Density Control \citep{desanti2025fdc} turns the local reward into
a distributional update. Starting from the pretrained distribution $p_0$, we
linearize both terms of the penalized objective at $p_k$ and constrain
the update with a KL-proximal penalty. The $k$-th subproblem is
\begin{equation}
p_{k+1}\in\arg\max_p
\left\{
\mathbb E_{X\sim p}\left[f_{n,p_k}(X) - \alpha \log \frac{p_k(X)}{p_0(X)}\right]
-\frac{1}{\eta_k}\mathrm{KL}(p\Vert p_k)
\right\},
\label{eq:proximal-update}
\end{equation}
The first variation of
$p\mapsto\KL(p\Vert p_0)$ at $p_k$ is
$\log(p_k(x)/p_0(x))+1$. The additive constant does not
affect the optimizer and is omitted in \cref{eq:proximal-update}.
Therefore, $f_{n,p_k}$ is the expected-hypervolume component of the
linearized reward, and the log-density ratio is the pretrained-model
penalty component. The final term,
$\eta_k^{-1}\KL(p\Vert p_k)$, is not part of the original objective but it is included to
limit how far the solution of the local subproblem can move from
$p_k$.

The linearized reward remains fixed during the $k$-th inner solve. Its
solution defines $p_{k+1}$, at which point we recompute both reward
components to form the next subproblem. Thus, the outer loop updates the
local reward, while the inner loop solves a standard scalar-reward
fine-tuning problem.

\paragraph{Theoretical guarantee.} The key question is whether progress
on successive local subproblems also improves the original penalized objective.
As shown in Appendix~\ref{app:per-candidate-coverage}, expected hypervolume
is concave in the sampling distribution, and the
pretrained-model KL penalty makes the objective strictly concave when
$\alpha>0$ (Appendix~\ref{app:regularization}). More importantly, relative smoothness bounds the error of
the linear approximation by a multiple of
$\KL(p_{k+1}\Vert p_k)$ (Appendix~\ref{app:relative-smoothness}). A sufficiently strong KL-proximal penalty
dominates this error, linking progress on each local subproblem to
progress on the original objective. The following theorem formalizes
this guarantee.

\begin{restatable}[Convergence of \AlgNameShort]{theorem}{hvftconvergence}
\label{thm:convergence}
Under the mild regularity conditions, the penalized objective in
\cref{eq:distributional-problem} admits a unique maximizer
$p^\star$ for $\alpha>0$. For any sufficiently small constant step
size $\eta$, the iterates in \cref{eq:proximal-update} are well defined,
monotonically improve the penalized objective, and converge
geometrically to $p^\star$. Precise conditions and rates are given in
Appendix~\ref{app:hvft-convergence}.
\end{restatable}

\paragraph{Practical implementation.} The expectation in
\cref{eq:variation-hvi} is approximated by Monte Carlo. At iteration
$k$, we draw independent background sets
$A_b\sim p_k^{\otimes(n-1)}$, for $b=1,\ldots,B$, and define
\begin{equation}
    \widehat f_{n,p_k}(x)
    :=
    \frac{n}{B}
    \sum_{b=1}^{B}
    \Delta_{\HV}(x;A_b).
    \label{eq:mc-reward}
\end{equation}
For every fixed $x$, this estimator is unbiased:
$\mathbb E[\widehat f_{n,p_k}(x)]=f_{n,p_k}(x)$.

During the $k$-th inner solve, we hold the background sets and their
objective values fixed, making $\widehat f_{n,p_k}$ a fixed empirical
reward. A model-specific scalar-reward method approximately solves the
subproblem over the generator parameters. After the outer iteration, we
draw new background sets from $p_{k+1}$ and construct the next reward. Our experiments in Section~\ref{sec:experiments} instantiate the inner solver with methods for continuous
flow-matching and discrete diffusion models
\citep{Zheng2025DiffusionNFTOD,tang2025tr2d2}.
Appendix~\ref{app:practicality} specifies the corresponding training
losses and treatment of local regularization.
\Cref{alg:expected-hv-finetuning} summarizes the procedure.

\begin{algorithm}[t]
\caption{\AlgNameDef}
\label{alg:expected-hv-finetuning}
\begin{algorithmic}[1]
\Require Pretrained distribution $p_0$, objective map $\rvec$, reference point $\rref$, set size $n\geq1$, background count $B\geq1$, penalty coefficient $\alpha\geq0$, maximum iterations $K$, step sizes $\eta_k>0$
\For{$k=0,\ldots,K-1$}
    \State Draw independent background sets $A_b\sim p_k^{\otimes(n-1)}$, $b=1,\ldots,B$
    \State Evaluate and cache $\rvec(a)$ for each background design $a\in A_b$
    \State Construct $\widehat f_{n,p_k}$ using \cref{eq:mc-reward}
    \State Approximately solve \cref{eq:proximal-update} with
    $f_{n,p_k}$ replaced by $\widehat f_{n,p_k}$
\EndFor
\State \Return $p_K$
\end{algorithmic}
\end{algorithm}

\section{Experiments}
\label{sec:experiments}

We evaluate \AlgNameShort on therapeutic-peptide and small-molecule design
tasks using discrete diffusion and continuous flow-matching models. Our
experiments test whether \AlgNameShort improves objective-space quality and
trade-off coverage while preserving validity and diversity in the
design space.

\paragraph{Evaluation protocol.}
At each evaluation, we generate candidates until obtaining 1,000 valid
designs and record the valid fraction over all generation attempts. From
this pool, we repeatedly draw random subsets of 10 and 100 observations
without replacement and report their mean hypervolume, while computing
diversity and design-space coverage over the full pool. We evaluate
100-hypervolume at regular intervals and use it to select the
best-performing epoch, applying the same rule to baselines wherever
possible. Across all tasks, \AlgNameShort uses $n=10$; we study the effect of $n$
in Appendix~\ref{app:ablation}. We repeat each experiment with three
random seeds, corresponding to independent fine-tuning runs for
training-based methods, and report means and standard errors across
runs. This across-run aggregation is distinct from the repeated subset
sampling used to estimate each run's hypervolume. Complete metric
definitions and evaluation details are provided in
Appendix~\ref{app:experimental-details}.

\subsection{Therapeutic peptides}

Peptides are molecules composed of amino-acid residues linked by peptide
bonds and are an established class of therapeutic agents
\citep{wang2022therapeutic}. We fine-tune the pretrained peptide masked
diffusion model of \citet{tang2025peptune} for five objectives: binding
affinity, solubility, non-hemolysis, non-fouling, and membrane
permeability. All objectives are evaluated using the classifiers and
regressors from the same work and are taken to be maximized.

We consider three binding targets: the transferrin receptor (TfR), a
common target for organ-specific drug delivery \citep{emily2025pep};
the glucagon-like peptide-1 receptor (GLP-1R), which is relevant to type
2 diabetes and obesity \citep{alfaris2024glp1}; and the
glutamate-aspartate transporter (GLAST), which is abundant on astrocytes
and implicated in neurological disorders \citep{pajarillo2019glast}.
Only peptides that pass the \verb|SMILES2PEPTIDE| verifier
\citep{tang2025peptune} are scored.

\paragraph{Baselines.}
Our primary fine-tuning baseline is \AlgNameTR
\citep{tang2025tr2d2}, a state-of-the-art method for multi-objective
fine-tuning of discrete diffusion models. We also include \AlgNameTR++, an
enhanced variant in which rewards are normalized to $[0,1]$ to avoid
scale bias and the best epoch is selected by hypervolume. For additional
context, we evaluate the pretrained model and \AlgNamePep
\citep{tang2025peptune}, an inference-time method that guides generation
using Monte Carlo tree search.

\paragraph{\AlgNameShort configuration.}
We use the single-objective variant of \AlgNameTR as \AlgNameShort's inner solver.
We leave the marginal hypervolume rewards unscaled because scaling gave
only minor improvements and requires a problem-dependent factor. See
Appendix~\ref{app:implementation} for details.

\begin{figure}[!h]
    \centering
    \includegraphics[width=1.0\linewidth]{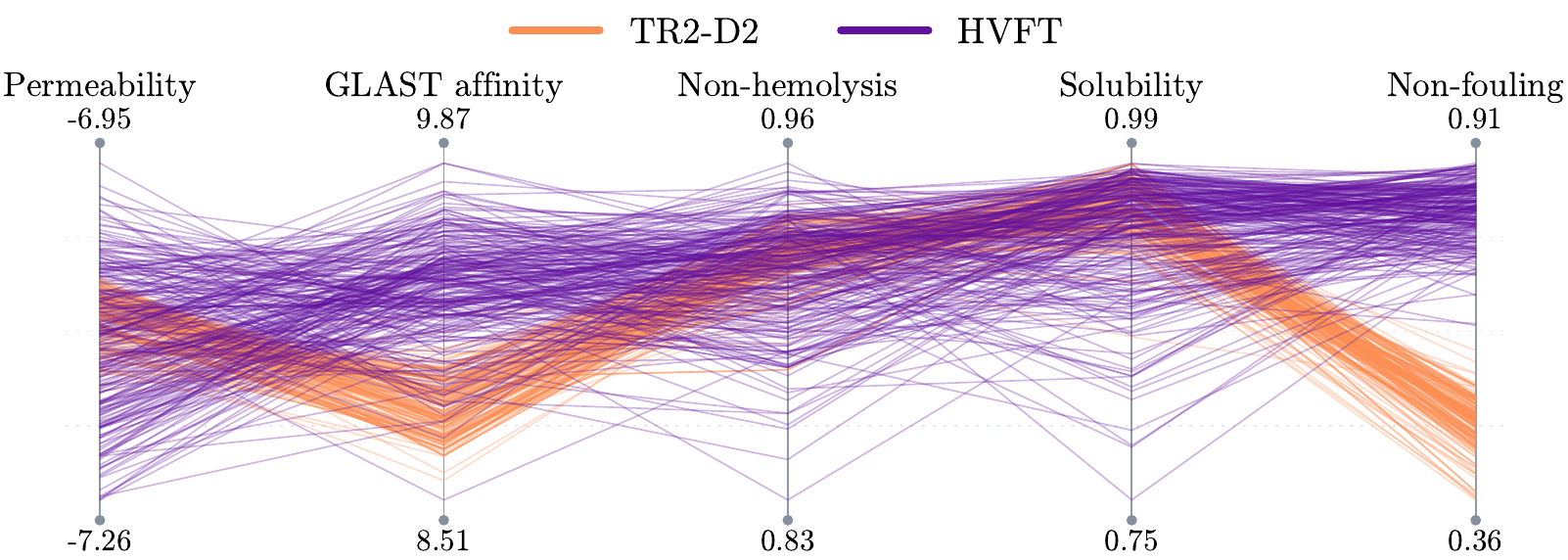}
    \caption{Pareto sets induced by 1,000 samples from \AlgNameTR \citep{tang2025tr2d2}, a state-of-the-art discrete diffusion fine-tuning algorithm, and \AlgNameShort, on the GLAST task. \AlgNameShort achieves broader objective-space coverage and higher individual sample quality than \AlgNameTR.
 }
    \label{fig:glast-pareto-comparison}
\vspace{-10pt}
\end{figure}

\paragraph{Results.}
As shown in Table~\ref{tab:peptide_results}, \AlgNameShort achieves the highest
10- and 100-hypervolume on all three targets. It also yields the highest
design-space diversity and coverage. Figure~\ref{fig:glast-pareto-comparison}
illustrates these gains on GLAST, where \AlgNameShort produces a Pareto set with
broader objective-space coverage and stronger individual candidates than
\AlgNameTR.

\begin{table*}[t]
\centering
\caption{\looseness -1 Objective-space and design-space metrics for three therapeutic peptide design tasks. Bold and underlined values indicate the highest and second-highest means, within each task. \AlgNameShort achieves the highest hypervolume across the three tasks, as well as the highest design-space diversity and coverage.}
\label{tab:peptide_results}
\setlength{\tabcolsep}{4pt}
\renewcommand{\arraystretch}{1.15}
\resizebox{\textwidth}{!}{%
\begin{tabular}{llccccc}
\hline
& & \multicolumn{2}{c}{Objective-space metrics}
  & \multicolumn{3}{c}{Design-space metrics} \\
\cline{3-4}\cline{5-7}
Task & Method & 100-Hypervolume & 10-Hypervolume
     & \shortstack{Diversity} & Coverage & Valid fraction \\
\hline

GLAST & Base
& $2.78 \pm 0.04$
& $1.61 \pm 0.01$
& $\underline{60.35 \pm 0.14}$
& $\underline{401.50 \pm 0.28}$
& $0.321 \pm 0.007$ \\

& \AlgNamePep
& $2.79 \pm 0.07$
& $1.59 \pm 0.03$
& $39.78 \pm 1.05$
& $134.95 \pm 5.87$
& $\mathbf{0.603 \pm 0.028}$ \\

& \AlgNameTR
& $7.78 \pm 1.06$
& $6.79 \pm 1.41$
& $9.08 \pm 6.24$
& $136.64 \pm 74.24$
& $0.249 \pm 0.031$ \\

& \AlgNameTR++
& $\underline{8.56 \pm 0.86}$
& $\underline{7.68 \pm 0.77}$
& $4.13 \pm 1.28$
& $89.43 \pm 28.69$
& $0.233 \pm 0.010$ \\

& \AlgNameShort
& $\mathbf{11.67 \pm 0.27}$
& $\mathbf{10.56 \pm 0.25}$
& $\mathbf{64.78 \pm 7.34}$
& $\mathbf{465.45 \pm 15.71}$
& $\underline{0.336 \pm 0.024}$ \\

\hline

GLP-1R & Base
& $3.01 \pm 0.04$
& $1.77 \pm 0.01$
& $\underline{60.35 \pm 0.14}$
& $\underline{401.50 \pm 0.28}$
& $0.321 \pm 0.007$ \\

& \AlgNamePep
& $3.02 \pm 0.11$
& $1.77 \pm 0.06$
& $40.46 \pm 1.06$
& $137.70 \pm 6.58$
& $\mathbf{0.614 \pm 0.012}$ \\

& \AlgNameTR
& $7.07 \pm 0.19$
& $5.64 \pm 0.12$
& $13.24 \pm 8.00$
& $182.56 \pm 77.61$
& $0.167 \pm 0.031$ \\

& \AlgNameTR++
& $\underline{8.93 \pm 0.98}$
& $\underline{8.12 \pm 0.83}$
& $3.59 \pm 0.82$
& $75.66 \pm 21.14$
& $\underline{0.349 \pm 0.060}$ \\

& \AlgNameShort
& $\mathbf{11.23 \pm 0.08}$
& $\mathbf{10.34 \pm 0.05}$
& $\mathbf{93.41 \pm 2.14}$
& $\mathbf{521.27 \pm 5.74}$
& $0.243 \pm 0.071$ \\

\hline

TfR & Base
& $2.93 \pm 0.04$
& $1.69 \pm 0.01$
& $\underline{60.35 \pm 0.14}$
& $\underline{401.50 \pm 0.28}$
& $0.321 \pm 0.007$ \\

& \AlgNamePep
& $2.82 \pm 0.18$
& $1.65 \pm 0.09$
& $39.97 \pm 0.78$
& $136.73 \pm 4.67$
& $\mathbf{0.551 \pm 0.059}$  \\

& \AlgNameTR
& $4.41 \pm 0.91$
& $2.78 \pm 1.14$
& $6.46 \pm 1.63$
& $126.79 \pm 34.38$
& $0.200 \pm 0.015$ \\

& \AlgNameTR++
& $\underline{7.63 \pm 1.87}$
& $\underline{6.75 \pm 1.90}$
& $4.91 \pm 1.45$
& $99.92 \pm 25.29$
& $\underline{0.331 \pm 0.068}$ \\

& \AlgNameShort
& $\mathbf{11.57 \pm 0.16}$
& $\mathbf{10.15 \pm 0.18}$
& $\mathbf{75.89 \pm 9.42}$
& $\mathbf{485.80 \pm 17.27}$
& $0.250 \pm 0.021$ \\

\hline
\end{tabular}%
}
\end{table*}

\subsection{Small molecules}

The vast space of drug-like molecules makes exhaustive screening
infeasible, motivating generative models that explore chemical space
and propose compounds with desirable properties
\citep{bombarelli2016automatic}. We fine-tune FlowMol
\citep{flomolDunn}, a continuous flow-matching model trained on
GEOM-Drugs \citep{Axelrod2020GEOMEM}, and evaluate two nested tasks:
\emph{Topology}, which uses graph-level objectives and validity checks,
and \emph{Geometry}, which adds conformation-level objectives and 3D
validity checks.

\begin{wrapfigure}{r}{0.43\textwidth}
    \centering
    \includegraphics[width=\linewidth]{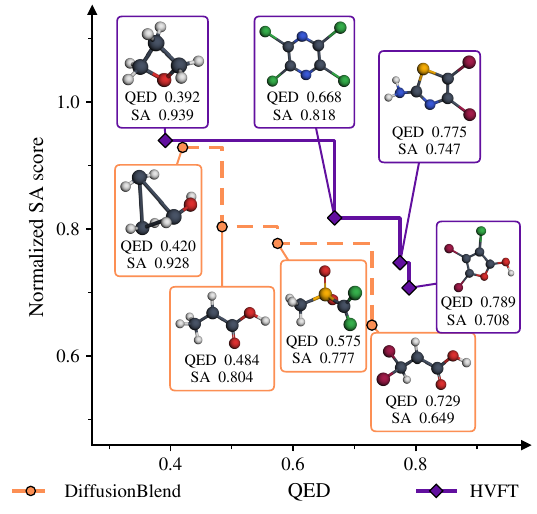}
    \vspace{-20pt}
    \caption{Best four-point subset by hypervolume, selected from 100 samples of DiffusionBlend and \AlgNameShort on the \emph{Topology} task. \AlgNameShort's samples achieve better coverage of high-quality trade-offs in objective-space.}
    \vspace{-15pt}
    \label{fig:diffusion_blend_vs_hvft}
\end{wrapfigure}
\looseness -1 For Topology, we maximize Quantitative Estimate of Drug-likeness (QED)
\citep{bickerton2012quantifying} and synthetic accessibility (SA)
\citep{SA2009Schu}, as implemented in RDKit
\citep{greg_landrum_2026_22140358}. We reverse and rescale the original
SA score to $[0,1]$ so that higher values are better. Only molecules that
pass the chemical validity checks of
\citet{Buttenschoen2025AnEO} are scored.

Geometry includes the Topology objectives and additionally maximizes
the dipole moment computed at the GFN2-xTB level of theory
\citep{bannwarth2019gfn2}. Scored molecules must also pass the
PoseBusters checks for bond lengths, bond angles, planar aromatic rings,
planar double bonds, and internal steric clashes
\citep{posebusetButten}.

\paragraph{Baselines.}
We compare \AlgNameShort against \AlgNameNFT \citep{Zheng2025DiffusionNFTOD}, a sample-efficient method that we apply
to a sum scalarization of the objectives, and \AlgNameBlend
\citep{Cheng2025DiffusionBI}, a state-of-the-art multi-objective method
that combines one separately aligned model per objective during sampling. Although its original implementation uses low-rank adapters,
we fully fine-tune each model with \AlgNameNFT to avoid restricting
model capacity. FlowMol is reported for context.

\paragraph{\AlgNameShort configuration.}
We use \AlgNameNFT as \AlgNameShort's scalar inner solver and keep its
hyperparameters fixed across \AlgNameShort and all fine-tuning baselines.
Details are provided in
Appendix~\ref{app:implementation}.

\paragraph{Results.}
Table~\ref{tab:molecular_results} shows that \AlgNameShort achieves the highest
10- and 100-hypervolume on both tasks. It also preserves substantially
more design-space diversity than the other fine-tuning
methods. Figure~\ref{fig:diffusion_blend_vs_hvft} compares representative
four-candidate subsets on Topology.

\begin{table}[]
  \centering
  \caption{Objective- and design-space metrics for the small-molecule design tasks. Bold and underlined values indicate the highest and second-highest means, respectively, within each task. \AlgNameShort achieves the highest hypervolume among state-of-the-art methods while preserving diversity and coverage.}
  \label{tab:molecular_results}
  \vspace{-5pt}
\setlength{\tabcolsep}{4pt}
\renewcommand{\arraystretch}{1.15}
\resizebox{\textwidth}{!}{%
    \begin{tabular}{llccccc}
      \hline
      & & \multicolumn{2}{c}{Objective-space metrics}
      & \multicolumn{3}{c}{Design-space metrics} \\
      \cline{3-4}\cline{5-7}
      Task & Method & 100-Hypervolume & 10-Hypervolume
             & Diversity & Coverage & Valid fraction \\
      \hline

      Topology & FlowMol
      & $3.116 \pm 0.014$
      & $2.665 \pm 0.007$
      & $\mathbf{266.1 \pm 4.9}$
      & $\mathbf{507.1 \pm 6.4}$
      & $0.705 \pm 0.008$ \\
      & \AlgNameNFT
      & $3.140 \pm 0.004$
      & $\underline{2.754 \pm 0.005}$
      & $128.8 \pm 2.7$
      & $312.8 \pm 8.7$
      & $\mathbf{0.952 \pm 0.009}$ \\
      & \AlgNameBlend
      & $\underline{3.337 \pm 0.024}$
      & $2.750 \pm 0.031$
      & $150.3 \pm 18.8$
      & $345.8 \pm 30.6$
      & $0.674 \pm 0.132$ \\
      & \AlgNameShort
      & $\mathbf{3.416 \pm 0.005}$
      & $\mathbf{3.247 \pm 0.004}$
      & $\underline{154.9 \pm 9.0}$
      & $\underline{351.6 \pm 11.1}$
      & $\underline{0.734 \pm 0.007}$ \\
      \hline
      Geometry & FlowMol
      & $4.279 \pm 0.024$
      & $3.655 \pm 0.010$
      & $\mathbf{91.4 \pm 2.3}$
      & $\mathbf{331.6 \pm 7.0}$
      & $0.483 \pm 0.009$ \\
      & \AlgNameNFT
      & $\underline{4.335 \pm 0.029}$
      & $3.669 \pm 0.018$
      & $70.9 \pm 1.6$
      & $\underline{269.4 \pm 6.8}$
      & $\underline{0.663 \pm 0.007}$ \\
      & \AlgNameBlend
      & $4.332 \pm 0.066$
      & $\underline{3.670 \pm 0.034}$
      & $51.6 \pm 1.9$
      & $187.4 \pm 2.2$
      & $\mathbf{0.707 \pm 0.043}$ \\
      & \AlgNameShort
      & $\mathbf{4.988 \pm 0.049}$
      & $\mathbf{4.151 \pm 0.024}$
      & $\underline{87.5 \pm 2.2}$
      & $249.8 \pm 1.9$
      & $0.518 \pm 0.017$ \\
      \hline
    \end{tabular}%
  }
  \vspace{-15pt}
\end{table}

\vspace{-3pt}
\subsection{Discussion}
\vspace{-3pt}

\AlgNameShort achieves the highest 10- and 100-hypervolume on every task,
showing consistent gains in candidate quality and collective trade-off
coverage across generative modalities and model architectures.
Crucially, \AlgNameShort builds on established methods as inner solvers:
single-objective \AlgNameTR for peptides and \AlgNameNFT for molecules.
Nevertheless, it outperforms \AlgNameTR and \AlgNameTR++ on the peptide tasks
and DiffusionNFT on the molecular tasks, supporting the benefit of the
sequential marginal-hypervolume rewards. \AlgNameShort also outperforms
DiffusionBlend and achieves these gains without manually specified
scalarization weights or preference allocation across objectives.

\looseness -1 Beyond objective-space performance, \AlgNameShort also performs strongly in the
design space. Among steering methods, it attains the highest diversity
on every task and the highest coverage on all but Geometry. In the peptide experiments, \AlgNameShort exceeds even the pretrained model in both metrics while maintaining a
comparable valid fraction. One possible explanation is that distinct
objective trade-offs may correspond to widely separated regions of the design space, so
seeking objective-space coverage can encourage broader design-space
exploration. Finally, the \AlgNamePep comparison places these fine-tuning gains in the
context of inference-time steering. The inference-time baseline achieves substantially lower hypervolume than the training-based methods on the peptide tasks. This pattern
is consistent with evidence that inference-time steering can have
limited ability to move generation beyond regions favored by the pretrained model
\citep{Klarner2024ContextGuidedDF,steinberg2025amortized,Santi2026ActiveFE}.

Despite strong performance, \AlgNameShort has limitations: it relies on a bi-level optimization structure and introduces hyperparameters (regularization $\alpha$, Monte-Carlo sample size $B$, set-size $n$; see Appendices~\ref{app:practicality} and \ref{app:ablation}). Additionally, hypervolume relies on the choice of reference point and is expensive to compute in many-objective settings. Monte-Carlo estimation can help reduce this cost, and extending \AlgNameShort to address many-objective scalability remains a direction for future work.

\vspace{-5pt}
\section{Conclusion}
\vspace{-5pt}
Multi-objective generative design requires samples that are individually
valuable and collectively span useful trade-offs, yet standard
fine-tuning relies on fixed per-candidate rewards that do not capture
this collective behavior. We introduced \AlgNameDef,
which adapts a pretrained generator to maximize expected hypervolume
while regularizing toward the pretrained distribution. Its key
observation is that the first variation of expected hypervolume is given
by the expected marginal contribution of a candidate and can be
estimated directly from model samples, yielding scalar-reward subproblems for which
we establish convergence guarantees.

Across therapeutic-peptide and small-molecule design tasks, \AlgNameShort
achieves the highest 10- and 100-hypervolume with both discrete
diffusion and continuous flow-matching models. It outperforms existing
multi-objective methods, including standalone versions of its inner
solvers, while preserving strong design-space diversity and coverage.
Together, these results establish \AlgNameShort as a principled and tractable
approach to fine-tuning generative models for high-quality,
complementary designs without predefined scalarization weights or
additional inference-time optimization.

\section*{Acknowledgments} 
This publication was made possible by the ETH AI Center doctoral fellowship to Riccardo De Santi.
The project has received funding from the Swiss
National Science Foundation under NCCR Catalysis (grant number 180544 and 225147) and NCCR Automation (grant agreement 51NF40 180545), National Centres of Competence in Research funded by the Swiss National Science Foundation.
\subsubsection*{Reproducibility statement}
The objective, first variation, estimator, and proofs are given in Section~\ref{sec:method} and Appendix~\ref{app:proofs}.
Section~\ref{sec:experiments} and Appendix~\ref{app:experimental-details} specify the evaluation rules and implementation choices.

\subsection*{AI use statement}
In this work, we used generative AI tools to identify related work, polish the writing, implement our algorithms in code, and assist with mathematical proofs.
All AI-assisted code was carefully checked and experimentally validated.
All AI-assisted proofs were independently verified.
All literature retrieved through AI was checked against the original sources.
We take full responsibility for the final content of this work, including all text, claims, and artifacts.

\bibliography{references}
\bibliographystyle{iclr2027_conference}

\appendix

\section{Theory}
\label{app:proofs}

The first two portions of these sections are dedicated to the expected-hypervolume objective $F_n(p)$ in Equation~\ref{eq:expected-hv}.
Appendix~\ref{app:first-variation} proves that the first variation of $F_n(p)$ is an expected marginal hypervolume contribution as defined in Proposition~\ref{thm:variation}.
This first variation, named $f_{n,p}$, drives the reward at the next iteration of HVFT.
Appendix~\ref{app:per-candidate-coverage} establishes concavity of our objective $F_n(p)$ in the sampling distribution and provides further intuition about the structure it induces on $p$.

Appendix~\ref{app:regularization} studies the regularized objective $F_n(p) - \alpha D_{\mathrm{KL}}(p \Vert p_0)$, and proves the existence of a unique maximizer $p^*$.
Appendix~\ref{app:relative-smoothness} then establishes curvature bounds of the regularized objective.
Equipped with these bounds, we prove the convergence guaranties of HVFT in Appendix~\ref{app:hvft-convergence}.

We use the notation of Section~\ref{sec:method} throughout.
We assume the objective map $\rvec$ is measurable, the reference point $\rref$ is fixed, and that $(r_m(x)-r_m^{\mathrm{ref}})_+\leq C$ for all $M$ objectives.
This final assumption ensures that $0\leq\HV(A;\rref)\leq C^M$ for every finite candidate set $A$.

\subsection{First variation of Expected Hypervolume}
\label{app:first-variation}

\marginalhvreward*

\begin{proof}
    Fix an integer $n\geq1$ and a probability distribution $p$ on
    $\mathcal X$.
    Let $h$ be a finite signed measure with $h(\mathcal X)=0$ such that
    $p_\epsilon=p+\epsilon h$ is a probability distribution for all
    sufficiently small $\epsilon\geq0$.
    Writing $x_{1:n}=(x_1,\ldots,x_n)$, the definition of
    expected hypervolume gives:
    \begin{align}
        F_n(p_\epsilon)
        &:=\mathbb E_{X_{1:n}\sim p_\epsilon^{\otimes n}}
        \left[
            \HV\!
            \left(\{X_1,\ldots,X_n\};\rref\right)
            \right] \nonumber \\
        &=\int_{\mathcal X^n}
        \HV(\{x_1,\ldots,x_n\};\rref)
        \,d\bigl(p_\epsilon^{\otimes n}\bigr)(x_{1:n}).
        \label{eq:app-perturbed-hv}
    \end{align}
    Let us recall the identity
    $(p_1+p_2)\otimes(q_1+q_2)
        =p_1\otimes q_1+p_1\otimes q_2+p_2\otimes q_1+p_2\otimes q_2$.
    Applying this identity repeatedly, we have:
    \begin{align}
        p_\epsilon^{\otimes n}
        &=(p+\epsilon h)^{\otimes n} \nonumber \\
        &=p^{\otimes n}
        +\epsilon\sum_{j=1}^n
        p^{\otimes(j-1)}\otimes h\otimes p^{\otimes(n-j)}
        + O(\epsilon^2).
        \label{eq:app-product-expansion}
    \end{align}

    Substituting \cref{eq:app-product-expansion} into
    \cref{eq:app-perturbed-hv} gives:
    \begin{align}
        F_n(p_\epsilon)
        & =\int_{\mathcal X^n}
        \HV(\{x_1,\ldots,x_n\};\rref)
        \,d\!
        \left[
            p^{\otimes n}+\epsilon\sum_{j=1}^n
            p^{\otimes(j-1)}\otimes h\otimes p^{\otimes(n-j)}
            \right](x_{1:n})
        +O(\epsilon^2)
        \nonumber \\
        & =\int_{\mathcal X^n}
        \HV(\{x_1,\ldots,x_n\};\rref)
        \,d(p^{\otimes n})(x_{1:n})
        \nonumber \\
        & \quad+\epsilon\int_{\mathcal X^n}
        \HV(\{x_1,\ldots,x_n\};\rref)
        \,d\!\left[\sum_{j=1}^n
            p^{\otimes(j-1)}\otimes h\otimes p^{\otimes(n-j)}
            \right](x_{1:n})
        +O(\epsilon^2)
        \nonumber \\
        & =F_n(p)
        +\epsilon\sum_{j=1}^n
        \int_{\mathcal X^n}
        \HV(\{x_1,\ldots,x_n\};\rref)
        \,d\bigl(p^{\otimes(j-1)}\otimes h\otimes p^{\otimes(n-j)}\bigr)(x_{1:n})
        +O(\epsilon^2).
        \label{eq:app-hv-integral-expansion}
    \end{align}
    In the first equality, the boundedness of hypervolume preserves
    the $O(\epsilon^2)$ remainder under integration.
    In the $j$th integral of \cref{eq:app-hv-integral-expansion},
    the coordinate $x_j$ is integrated against $h$, while every other
    coordinate is integrated against $p$.
    Since hypervolume is symmetrical, we can change the order of integration to have $x_j$ as the final integral.
    Thus every term inside the summand is equal and we can write the following expression:
    \begin{equation}
        F_n(p_\epsilon)=F_n(p)
        +\epsilon n\int_{\mathcal X}
        \mathbb E_{A\sim p^{\otimes(n-1)}}
        \left[\HV(A\cup\{x\};\rref)\right]\,dh(x)
        +O(\epsilon^2).
        \label{eq:app-hv-linear-term}
    \end{equation}
    Here $x$ denotes the candidate integrated against $h$, and $A$ is the
    set formed by the other $n-1$ independent draws from $p$.
    We are almost done.
    We just need to rewrite \cref{eq:app-hv-linear-term} in terms of the marginal hypervolume contribution in \cref{eq:marginal-hv}.
    To do this, we note that the background hypervolume is independent of $x$, and thus:
    \begin{equation}
        \int_{\mathcal X}
        \mathbb E_{A\sim p^{\otimes(n-1)}}
            [\HV(A;\rref)]\,dh(x)
        =\mathbb E_A[\HV(A;\rref)]\,h(\mathcal X)=0.
        \label{eq:app-equal-to-zero}
    \end{equation}
    We may therefore subtract (\ref{eq:app-equal-to-zero}) inside the linear term in
    (\ref{eq:app-hv-linear-term}):
    \begin{align}
        & n\int_{\mathcal X}
        \mathbb E_A[\HV(A\cup\{x\};\rref)]\,dh(x)
        \nonumber                                        \\
        & \qquad=n\int_{\mathcal X}
        \mathbb E_A\left[
            \HV(A\cup\{x\};\rref)-\HV(A;\rref)
            \right]\,dh(x)
        \nonumber                                        \\
        & \qquad=n\int_{\mathcal X}
        \mathbb E_A\left[\Delta_{\HV}(x;A)\right]\,dh(x)
        \qquad\text{(Definition in Equation~\ref{eq:marginal-hv})}
        \nonumber                                        \\
        & \qquad=\int_{\mathcal X}f_{n,p}(x)\,dh(x).
        \qquad\text{(Definition in Equation~\ref{eq:variation-hvi})}
    \end{align}
    Consequently,
    \begin{align*}
        F_n(p_\epsilon)
        &=F_n(p)+\epsilon\int_{\mathcal X}f_{n,p}(x)\,dh(x)
        +O(\epsilon^2) \\
        &=F_n(p)+ \epsilon \mathbb{E}_{h} \left[ f_{n,p}(X)\right]
        +o(\epsilon)
    \end{align*}

    Finally, for any admissible probability distribution $q$, taking
    $h=q-p$ gives $p_\epsilon=(1-\epsilon)p+\epsilon q$ and
    \begin{align*}
        F_n\left((1-\epsilon)p+\epsilon q\right)
        = F_n(p_\epsilon) &= F_n(p)+ \epsilon\mathbb{E}_{h} \left[ f_{n,p}(X)\right]
        +o(\epsilon) \\
        & = F_n(p)
        +
        \epsilon\left(
        \mathbb E_q[f_{n,p}(X)]
        -
        \mathbb E_p[f_{n,p}(X)]
        \right)
        +
        o(\epsilon)
    \end{align*}
    This yields the first-variation identity in
    \cref{eq:first-variation}.
\end{proof}

\subsection{Concavity of Expected Hypervolume} \label{app:per-candidate-coverage}
Define $\phi_n(s):=1-(1-s)^n$ for $s\in[0,1]$.
This is the probability of at least one success in $n$ independent
trials, each with success probability $s$.
In this section, we derive an alternative representation of $F_n(p)$ in terms of \emph{coverage}, defined shortly.
This representation will make concavity apparent.

For a fixed objective-space point $u\succeq\rref$ and design $X \in \mathcal{X}$, we say that $x$ \emph{covers} $u$ if and only if $\rvec(x) \succeq u$.
When $x$ is sampled from a distribution $p$, $\rvec(x) \succeq u$ is a probabilistic event.
This leads us to the definition of the coverage probability $S_p(u)$:
\begin{equation}
    S_p(u):=\Pr_{X\sim p}\!
    \left(\rvec(X)\succeq u\right)
    \label{eq:app-coverage-probability}
\end{equation}
A set $A$ covers $u$ if at least one of its elements covers $u$.
For $A=\{X_1,\ldots,X_n\}$ with $X_1,\ldots,X_n\overset{\mathrm{i.i.d.
        }}{\sim}p$,
independence implies that all $n$ draws fail to cover $u$ with probability
$(1-S_p(u))^n$, hence
\[
    \Pr\!\left(\exists x\in A:\rvec(x)\succeq u\right)
    =1-(1-S_p(u))^n = \phi_n(S_p(u)).
\]

We can rewrite hypervolume in term of coverage, as follows:
\begin{align*}
    \HV(A;\rref)
    &= \operatorname{Leb}_M
    \left(
    \bigcup_{x\in A}
    \left\{
    u\in\mathbb R^M:
    \rref\preceq u\preceq\rvec(x)
    \right\}
    \right) \\
    &= \operatorname{Leb}_M
    \left(
    \bigcup_{x\in A}
    \bigcup_{u\in \mathbb{R}^M}
    \left\{
    \rref\preceq u\preceq\rvec(x)
    \right\}
    \right) \\
    &=\operatorname{Leb}_M
    \left(
    \bigcup_{u\in \mathbb{R}^M}
    \bigcup_{x\in A}
    \left\{
    \rref\preceq u\preceq\rvec(x)
    \right\}
    \right)\\
    &=\int_{\mathbb{R}^M}
    \ind\{\exists x\in A:\rvec(x)\succeq u\}\,du \\
    &=\int_{u \succeq \rref}
    \ind\{\exists x\in A:\rvec(x)\succeq u\}\,du.
\end{align*}

Taking expectations under the product measure $p^{\otimes n}$,
applying Tonelli's theorem, and using the coverage probability above gives
\begin{align}
    F_n(p)
    & :=\mathbb E_{X_{1:n}\sim p^{\otimes n}}
    \left[\HV\!
        \left(\{X_1,\ldots,X_n\};\rref\right)\right]
    \nonumber \\
    & =\int_{u\succeq\rref}
    \mathbb E_{X_{1:n}\sim p^{\otimes n}}
    \left[\ind\{\exists i\leq n:\rvec(X_i)\succeq u\}\right]\,du
    \nonumber \\
    & =\int_{u\succeq\rref}
    \Pr\!
    \left(\exists i\leq n:\rvec(X_i)\succeq u\right)\,du
    \nonumber                \\
    & =\int_{u\succeq\rref}
    \left[1-(1-S_p(u))^n\right]\,du
    \nonumber \\
    & =\int_{u\succeq\rref}\phi_n(S_p(u))\,du.
    \label{eq:coverage}
\end{align}

From the representation in Equation~\ref{eq:coverage}, we can proceed to show concavity of $F_n(p)$
\begin{theorem}[Concavity of expected hypervolume]
    \label{thm:coverage}
    For every integer $n\geq1$, $F_n$ is concave on
    $\mathcal P(\mathcal X)$.
    In particular, for any probability
    distributions $p,q$ and $t\in[0,1]$,
    \begin{equation}
        F_n((1-t)p+tq)\geq(1-t)F_n(p)+tF_n(q).
        \label{eq:app-hv-concavity}
    \end{equation}
\end{theorem}

\begin{proof}

    For each fixed $u$, the quantity $S_p(u)$ is linear in
    the distribution $p$, since it is just the expectation $\mathbb E _p$ of the fixed indicator function $\ind\left[{\rvec(X) \succeq u}\right]$.
    Meanwhile, $\phi_n$ is increasing and
    concave on $[0,1]$.
    Thus, $\phi_n(S_p(u))$ is concave in $p$, and
    integration preserves concavity.
\end{proof}

The coverage representation in Equation~\ref{eq:coverage} clarifies the preferences induced by $F_n$.
The marginal value of increasing a point's coverage probability is
$\phi_n'(s)=n(1-s)^{n-1}$, which decreases in $s$ for $n>1$.
Thus, the same increase in coverage probability yields a greater benefit
at a point that is rarely covered than at one that is already reached
reliably.
This diminishing-returns structure favors allocating probability
mass across complementary trade-offs, without requiring a fixed set of
preference directions to be specified in advance.

\subsection{KL regularization, existence, and uniqueness}\label{app:regularization}
HVFT optimizes expected hypervolume while staying close to a prior, as shown in the optimization problem (\ref{eq:distributional-problem}).
We show this regularized objective has a unique maxizer $p^*$.
For $\alpha>0$, write the objective in
\cref{eq:distributional-problem} as
\begin{equation}
    G_{n,\alpha}(p):=F_n(p)-\alpha\KL(p\Vert p_0),
    \qquad
    p\in\mathcal D:=\{p\in\mathcal P(\mathcal X):
    \KL(p\Vert p_0)<\infty\}.
    \label{eq:app-regularized-objective}
\end{equation}
KL divergence is strictly convex in its first argument on the convex
domain $\mathcal D$, so $-\alpha\KL(p\Vert p_0)$ is strictly
concave \citep{desanti2025fdc}.
Because summation preserves (strict) concavity and $F_n(p)$ is concave (see Theorem~\ref{thm:coverage}), $G_{n,\alpha}$ is also strictly concave.

A consequence of strict concavity is that if $G_{n,\alpha}$ attains its maximum on $\mathcal D$, its maximizer $p^\star$ is unique.
To see this, assume for the sake of contradiction that there exist two maximizers $p$ and $q$.
Strict concavity would give $G_{n,\alpha}((p+q)/2)>(G_{n,\alpha}(p)+G_{n,\alpha}(q))/2$, contradicting maximality.

The KL penalty also provides the control needed to prove existence.
We show that distributions whose objective values approach the supremum
must converge to a common limit, and that this limit attains the
supremum.

\begin{theorem}[Existence of a regularized maximizer]
    \label{thm:app-existence}
    Under the standing measurability and boundedness assumptions on
    $\rvec$, for every integer $n\geq1$ and $\alpha>0$, $G_{n,\alpha}$
    attains its maximum on $\mathcal D$.
\end{theorem}
\begin{proof}
    Let $S:=\sup_{q\in\mathcal D}G_{n,\alpha}(q)$.
    Since $p_0\in\mathcal D$, $\KL(p_0\Vert p_0)=0$, and
    $0\leq F_n\leq C^M$, we have
    $G_{n,\alpha}(p_0)\leq S\leq C^M$.
    Choose a maximizing sequence $(p_j)\subset\mathcal D$ with
    $S-1/j<G_{n,\alpha}(p_j)\leq S$.
    Its KL divergences satisfy the uniform bound
    \[
        \KL(p_j\Vert p_0)
        =\frac{F_n(p_j)-G_{n,\alpha}(p_j)}{\alpha}
        \leq\frac{C^M-G_{n,\alpha}(p_0)+1}{\alpha}
        =:B<\infty.
    \]
    Finite KL divergence implies $p_j\ll p_0$.
    By the Radon--Nikodym theorem, each $p_j$ therefore has a density
    $v_j:=dp_j/dp_0$ relative to $p_0$: for every measurable
    $A\subseteq\mathcal X$,
    \[
        p_j(A)=\int_A v_j\,dp_0,
        \qquad v_j\geq0\quad p_0\text{-almost everywhere},
        \qquad \int v_j\,dp_0=1.
    \]
    In particular, these Radon--Nikodym derivatives belong to $L^1(p_0)$.

    We next show that these densities form a Cauchy sequence.
    Since $(p_j+p_k)/2\in\mathcal D$, the entropy midpoint identity gives
    \begin{align*}
        &\frac{\KL(p_j\Vert p_0)+\KL(p_k\Vert p_0)}{2}
        -\KL\!\left(\frac{p_j+p_k}{2}\middle\Vert p_0\right) \\
        &\qquad=\frac12\left[
            \KL\!\left(p_j\middle\Vert\frac{p_j+p_k}{2}\right)
            +\KL\!\left(p_k\middle\Vert\frac{p_j+p_k}{2}\right)
        \right].
    \end{align*}
    Combining this identity with concavity of $F_n$ gives
    \begin{align*}
        G_{n,\alpha}\!\left(\frac{p_j+p_k}{2}\right)
        &\geq\frac{G_{n,\alpha}(p_j)+G_{n,\alpha}(p_k)}{2} \\
        &\quad+\frac{\alpha}{2}\left[
            \KL\!\left(p_j\middle\Vert\frac{p_j+p_k}{2}\right)
            +\KL\!\left(p_k\middle\Vert\frac{p_j+p_k}{2}\right)
        \right].
    \end{align*}
    Pinsker's inequality bounds each KL divergence below by half the
    squared $L^1(p_0)$ distance between the corresponding densities.
    The density of $(p_j+p_k)/2$ relative to $p_0$ is $(v_j+v_k)/2$, so
    \begin{align*}
        \KL\!\left(p_j\middle\Vert\frac{p_j+p_k}{2}\right)
        &\geq\frac12\left\|v_j-\frac{v_j+v_k}{2}\right\|_{L^1(p_0)}^2
        =\frac18\|v_j-v_k\|_{L^1(p_0)}^2, \\
        \KL\!\left(p_k\middle\Vert\frac{p_j+p_k}{2}\right)
        &\geq\frac12\left\|v_k-\frac{v_j+v_k}{2}\right\|_{L^1(p_0)}^2
        =\frac18\|v_j-v_k\|_{L^1(p_0)}^2.
    \end{align*}
    Substituting these bounds into the preceding inequality yields
    \begin{align*}
        G_{n,\alpha}\!\left(\frac{p_j+p_k}{2}\right)
        &\geq\frac{G_{n,\alpha}(p_j)+G_{n,\alpha}(p_k)}{2} \\
        &\quad+\frac{\alpha}{2}\left(
            \frac18\|v_j-v_k\|_{L^1(p_0)}^2
            +\frac18\|v_j-v_k\|_{L^1(p_0)}^2\right) \\
        &=\frac{G_{n,\alpha}(p_j)+G_{n,\alpha}(p_k)}{2}
        +\frac{\alpha}{8}\|v_j-v_k\|_{L^1(p_0)}^2.
    \end{align*}
    Since $G_{n,\alpha}((p_j+p_k)/2)\leq S$, it follows that
    \[
        \|v_j-v_k\|_{L^1(p_0)}^2
        \leq\frac{4}{\alpha}
        \left(2S-G_{n,\alpha}(p_j)-G_{n,\alpha}(p_k)\right)
        \leq\frac{4}{\alpha}\left(\frac1j+\frac1k\right).
    \]
    Completeness of $L^1(p_0)$ therefore gives a limit
    $v_j\to v$ in $L^1(p_0)$.
    Nonnegativity and unit integral pass to this limit, so
    $p(A):=\int_A v\,dp_0$ defines a probability distribution.

    We have shown that $p$ is a probability distribution.
    To establish $p\in\mathcal D$, it therefore remains only to show
    that $\KL(p\Vert p_0)<\infty$.
    The bounded scalar sequence $(\KL(p_j\Vert p_0))_{j\geq1}$ has a
    subsequence converging to its limit inferior.
    The corresponding densities still converge to $v$ in $L^1(p_0)$,
    so they admit a further subsequence converging $p_0$-almost everywhere.
    Denote the indices of this final subsequence of the original
    maximizing sequence by $j_k$. Thus both
    \[
        v_{j_k}\longrightarrow v\quad p_0\text{-almost everywhere},
        \qquad
        \KL(p_{j_k}\Vert p_0)
        \longrightarrow\liminf_{j\to\infty}\KL(p_j\Vert p_0).
    \]
    With the convention $0\log0=0$, Fatou's lemma applied to the
    nonnegative functions $v_{j_k}\log v_{j_k}+1/e$ gives
    \begin{align*}
        \KL(p\Vert p_0)
        =\int v\log v\,dp_0
        &\leq\liminf_{k\to\infty}
            \int v_{j_k}\log v_{j_k}\,dp_0 \\
        &=\liminf_{j\to\infty}\KL(p_j\Vert p_0)
        \leq B<\infty.
    \end{align*}
    Thus $p\in\mathcal D$.

    Finally, we show that $p$ attains $S$.
    For the full sequence $(p_j)$ and every objective-space point $u$,
    \[
        |S_{p_j}(u)-S_p(u)|
        =\left|\int\ind\{\rvec(x)\succeq u\}
            (v_j(x)-v(x))\,dp_0(x)\right|
        \leq\|v_j-v\|_{L^1(p_0)}\longrightarrow0.
    \]
    The integrands in the coverage representation~\cref{eq:coverage}
    therefore converge pointwise. They lie in $[0,1]$ and vanish
    outside the finite coverage box, so dominated convergence gives
    $F_n(p_j)\to F_n(p)$.
    Combining this convergence with the lower semicontinuity of KL
    established above yields
    \[
        G_{n,\alpha}(p)
        \geq\limsup_jG_{n,\alpha}(p_j)=S.
    \]
    Since $p\in\mathcal D$, the reverse inequality follows from the
    definition of $S$. Hence $G_{n,\alpha}(p)=S$.
\end{proof}

\paragraph{Observation: first variation of the regularized objective.}
As in the main text, write $p(x)$ and $p_0(x)$ for densities with
respect to a common dominating measure.
When $p$ and $p_0$ are
mutually absolutely continuous and $\log(p/p_0)$ is bounded
$p_0$-almost everywhere, a first variation of $G_{n,\alpha}$ is
\begin{equation}
    g_{\alpha,p}(x)
    :=f_{n,p}(x)-\alpha\log\frac{p(x)}{p_0(x)}.
    \label{eq:app-full-reward}
\end{equation}
Indeed, the first variation of $\KL(p\Vert p_0)$ is
$\log(p(x)/p_0(x))+1$.
The constant $-\alpha$ can be omitted
because it integrates to zero against $q-p$.
Thus, the full
linearized reward combines the marginal hypervolume reward with
the log-ratio penalty used in \cref{eq:proximal-update}.

\subsection{Bounding Linear Approximation Error: Relative Smoothness}
\label{app:relative-smoothness}

We bound the error of the linear approximations used by HVFT in
terms of KL divergence.
For a concave functional, this means bounding
how far the functional can fall below its tangent approximation;
equivalently, its negative is smooth relative to entropy.

\subsubsection{Relative smoothness of $F_n(p)$}
\label{app:fn-relative-smoothness}

Define
\[
    L_n:=\frac{n(n-1)C^M}{4},
    \qquad
    \|q-p\|_{\mathrm{TV}}:=\sup_{E\subseteq\mathcal X\text{ measurable}}
    |q(E)-p(E)|.
\]

\begin{proposition}[Relative smoothness of expected hypervolume]
    \label{prop:app-fn-smoothness}
    For every integer $n\geq1$ and probability distributions $p,q$ with
    $\KL(q\Vert p)<\infty$,
    \begin{equation}
        0\leq F_n(p)+\int_{\mathcal X}f_{n,p}(x)\,d(q-p)(x)-F_n(q)
        \leq L_n\KL(q\Vert p).
        \label{eq:app-fn-smoothness}
    \end{equation}
\end{proposition}
\begin{proof}
    For $n\geq2$, set $p_t=(1-t)p+tq$ and $\psi(t)=F_n(p_t)$.
    Since $S_{p_t}(u)=(1-t)S_p(u)+tS_q(u)$, differentiating the
    coverage representation in \cref{eq:coverage} twice gives
    \[
        \psi''(t)=-n(n-1)\int_{u\succeq\rref}
        (1-S_{p_t}(u))^{n-2}(S_q(u)-S_p(u))^2\,du.
    \]
    The finite coverage box and bounded integrands justify differentiation
    under the integral; endpoint values are understood by continuity.
    Because $S_q(u)-S_p(u)$ is the difference of the probabilities of
    the same measurable event,
    $|S_q(u)-S_p(u)|\leq\|q-p\|_{\mathrm{TV}}$.
    Thus
    \[
        -n(n-1)C^M\|q-p\|_{\mathrm{TV}}^2\leq\psi''(t)\leq0.
    \]
    Taylor's formula with integral remainder, together with
    $\psi'(0)=\int f_{n,p}(x)\,d(q-p)(x)$, implies
    \begin{align*}
        0\leq\psi(0)+\psi'(0)-\psi(1)
        & =-\int_0^1(1-t)\psi''(t)\,dt                   \\
        & \leq\frac{n(n-1)C^M}{2}\|q-p\|_{\mathrm{TV}}^2
        \leq L_n\KL(q\Vert p),
    \end{align*}
    where the last inequality is Pinsker's inequality.
    For $n=1$,
    $F_1$ is linear and the result holds with $L_1=0$.
\end{proof}

\subsubsection{Relative smoothness of $G_{n,\alpha}(p)$}
\label{app:ga-relative-smoothness}

The KL regularizer also contributes to the linearization error.
Using the first variation in \cref{eq:app-full-reward}, we account
for the log-ratio term explicitly.
Its contribution to the error is
exactly $\alpha\KL(q\Vert p)$, which adds to the relative smoothness
constant and supplies a quantitative strong-concavity bound.

\begin{proposition}[Relative smoothness and strong concavity of $G_{n,\alpha}$]
    \label{prop:app-ga-curvature}
    Let $\alpha>0$, $p,q\in\mathcal D$, and suppose that $p$ is
    equivalent to $p_0$ and $\log(p/p_0)\in L^\infty(p_0)$.
    Then
    \begin{align}
        \alpha\KL(q\Vert p)
        & \leq G_{n,\alpha}(p)
        +\int_{\mathcal X}g_{\alpha,p}(x)\,d(q-p)(x)-G_{n,\alpha}(q)
        \nonumber                         \\
        & \leq(L_n+\alpha)\KL(q\Vert p).
        \label{eq:app-ga-curvature}
    \end{align}
    Thus $-G_{n,\alpha}$ is $(L_n+\alpha)$-smooth relative to entropy,
    and $G_{n,\alpha}$ is $\alpha$-strongly concave relative to entropy
    for these pairs of distributions.
\end{proposition}
\begin{proof}
    The assumptions ensure that all terms in the entropy identity
    \[
        \KL(q\Vert p_0)-\KL(p\Vert p_0)
        -\int_{\mathcal X}\log\frac{p(x)}{p_0(x)}\,d(q-p)(x)
        =\KL(q\Vert p)
    \]
    are finite.
    Substituting
    $G_{n,\alpha}(p)=F_n(p)-\alpha\KL(p\Vert p_0)$ and
    $g_{\alpha,p}(x)=f_{n,p}(x)-\alpha\log(p(x)/p_0(x))$ into the linearization error
    gives
    \begin{align*}
        & G_{n,\alpha}(p)
        +\int_{\mathcal X}g_{\alpha,p}(x)\,d(q-p)(x)-G_{n,\alpha}(q) \\
        & \quad=F_n(p)-\alpha\KL(p\Vert p_0)
        +\int_{\mathcal X}\bigl[f_{n,p}(x)-\alpha\log\frac{p(x)}{p_0(x)}\bigr]
        \,d(q-p)(x)                                                  \\
        & \qquad\quad-F_n(q)+\alpha\KL(q\Vert p_0)                  \\
        & \quad=\left[F_n(p)
        +\int_{\mathcal X}f_{n,p}(x)\,d(q-p)(x)-F_n(q)\right]        \\
        & \qquad\quad+\alpha\left[\KL(q\Vert p_0)-\KL(p\Vert p_0)
            -\int_{\mathcal X}\log\frac{p(x)}{p_0(x)}\,d(q-p)(x)\right]  \\
        & \quad=\left[F_n(p)
        +\int_{\mathcal X}f_{n,p}(x)\,d(q-p)(x)-F_n(q)\right]
        +\alpha\KL(q\Vert p).
    \end{align*}
    The last equality uses the KL identity above.
    By
    Proposition~\ref{prop:app-fn-smoothness}, the bracketed
    linearization error of $F_n$ lies between $0$ and
    $L_n\KL(q\Vert p)$.
    Adding $\alpha\KL(q\Vert p)$ to these
    two bounds gives \cref{eq:app-ga-curvature}: the lower bound
    establishes relative strong concavity, and the upper bound
    establishes relative smoothness.
\end{proof}

\subsection{Convergence of the linearized HVFT update}
\label{app:hvft-convergence}

We now analyze the exact distributional update in
\cref{eq:proximal-update}, which uses the full reward
$g_{\alpha,p_k}$, including the linearized log-ratio penalty:
\begin{equation}
    p_{k+1}=\arg\max_{p\in\mathcal D}
    \left\{\mathbb E_p[g_{\alpha,p_k}(X)]
    -\eta^{-1}\KL(p\Vert p_k)\right\}.
    \label{eq:app-linearized-update}
\end{equation}
The pretrained-model penalty enters through
$-\alpha\log(p_k(x)/p_0(x))$ in the reward.
Its curvature is accounted
for by the $L_n+\alpha$ step-size bound below.

Throughout the next three lemmas, assume the standing measurability
and boundedness conditions, fix $n\geq1$ and $\alpha>0$, initialize
at $p_0$, and choose
$0<\eta\leq(L_n+\alpha)^{-1}$.
Write $\rho=1-\alpha\eta\in[0,1)$.

\begin{lemma}[Well-defined updates]
    \label{lem:app-well-defined-updates}
    Every subproblem in \cref{eq:app-linearized-update} has a unique
    solution $p_{k+1}\in\mathcal D$, given by
    \begin{equation}
        p_{k+1}(x)
        =\frac{p_k(x)\exp(\eta g_{\alpha,p_k}(x))}{Z_k},
        \qquad
        Z_k:=\mathbb E_{p_k}\!
        \left[\exp(\eta g_{\alpha,p_k}(X))\right].
        \label{eq:app-linearized-tilt}
    \end{equation}
    Each iterate is mutually absolutely continuous with $p_0$, and
    $\log(p_k/p_0)$ is bounded $p_0$-almost everywhere.
\end{lemma}
\begin{proof}
    Initially $\log(p_0/p_0)=0$.
    If $\log(p_k/p_0)$ is bounded,
    then $g_{\alpha,p_k}$ is bounded because
    $0\leq f_{n,p_k}\leq nC^M$.
    Consequently $0<Z_k<\infty$, and
    the exponential update defines a probability distribution with the
    same null sets as $p_k$.
    Moreover,
    \[
        \log\frac{p_{k+1}(x)}{p_0(x)}
        =\log\frac{p_k(x)}{p_0(x)}
        +\eta g_{\alpha,p_k}(x)-\log Z_k
    \]
    is bounded.
    Induction therefore gives $p_{k+1}\in\mathcal D$
    and the conditions required by Proposition~\ref{prop:app-ga-curvature}
    at every iteration.

    For any $p\in\mathcal D$, substituting the exponential update
    into $\KL(p\Vert p_{k+1})$ and rearranging yields
    \[
        \mathbb E_p[g_{\alpha,p_k}(X)]-\eta^{-1}\KL(p\Vert p_k)
        =\eta^{-1}\log Z_k-\eta^{-1}\KL(p\Vert p_{k+1}).
    \]
    All terms are finite by the bounded log ratios and $p\in\mathcal D$.
    Since KL divergence is nonnegative and vanishes exactly at
    $p=p_{k+1}$, this distribution uniquely solves the subproblem.
\end{proof}

\begin{lemma}[One-step bound and monotonic improvement]
    \label{lem:app-one-step-bound}
    For every $q\in\mathcal D$ and $k\geq0$, the iterates satisfy
    \begin{align}
        G_{n,\alpha}(q)-G_{n,\alpha}(p_{k+1})
        & \leq(\eta^{-1}-\alpha)\KL(q\Vert p_k)
        -\eta^{-1}\KL(q\Vert p_{k+1})
        \nonumber                                 \\
        & \quad-\bigl(\eta^{-1}-L_n-\alpha\bigr)
        \KL(p_{k+1}\Vert p_k).
        \label{eq:app-linearized-master-bound}
    \end{align}
    In particular, $G_{n,\alpha}(p_{k+1})\geq G_{n,\alpha}(p_k)$.
\end{lemma}
\begin{proof}
    Taking logarithms of the update in
    Lemma~\ref{lem:app-well-defined-updates} gives
    \[
        \eta g_{\alpha,p_k}(x)
        =\log\frac{p_{k+1}(x)}{p_k(x)}+\log Z_k.
    \]
    Integrate against $q-p_{k+1}$.
    The constant $\log Z_k$ cancels
    because both distributions have mass one.
    Furthermore,
    \[
        \int_{\mathcal X}\log\frac{p_{k+1}(x)}{p_k(x)}\,dq(x)
        =\KL(q\Vert p_k)-\KL(q\Vert p_{k+1}),
    \]
    while integration of the same logarithm against $p_{k+1}$ gives
    $\KL(p_{k+1}\Vert p_k)$.
    Therefore
    \begin{align}
        \int_{\mathcal X}g_{\alpha,p_k}(x)\,d(q-p_{k+1})(x)
        =\eta^{-1}\bigl[ & \KL(q\Vert p_k)-\KL(q\Vert p_{k+1})
            \nonumber                                              \\
            & -\KL(p_{k+1}\Vert p_k)\bigr].
        \label{eq:app-linearized-three-point}
    \end{align}

    Next apply the two bounds in \cref{eq:app-ga-curvature} at the
    same base point $p_k$.
    Relative strong concavity bounds the
    comparator's objective from above:
    \[
        G_{n,\alpha}(q)\leq G_{n,\alpha}(p_k)
        +\int_{\mathcal X}g_{\alpha,p_k}(x)\,d(q-p_k)(x)
        -\alpha\KL(q\Vert p_k).
    \]
    Relative smoothness bounds the next iterate's objective from below:
    \[
        G_{n,\alpha}(p_{k+1})\geq G_{n,\alpha}(p_k)
        +\int_{\mathcal X}g_{\alpha,p_k}(x)\,d(p_{k+1}-p_k)(x)
        -(L_n+\alpha)\KL(p_{k+1}\Vert p_k).
    \]
    Subtracting these inequalities cancels $G_{n,\alpha}(p_k)$ and combines
    the two integrals into an integral against $q-p_{k+1}$:
    \begin{align*}
        G_{n,\alpha}(q)-G_{n,\alpha}(p_{k+1})
        & \leq\int_{\mathcal X}g_{\alpha,p_k}(x)\,d(q-p_{k+1})(x)
        -\alpha\KL(q\Vert p_k)                                     \\
        & \quad+(L_n+\alpha)\KL(p_{k+1}\Vert p_k).
    \end{align*}
    Substituting \cref{eq:app-linearized-three-point} and collecting
    the coefficients of each KL divergence proves
    \cref{eq:app-linearized-master-bound}.
    For $q=p_k$, it gives
    \[
        G_{n,\alpha}(p_{k+1})-G_{n,\alpha}(p_k)
        \geq\eta^{-1}\KL(p_k\Vert p_{k+1})
        +(\eta^{-1}-L_n-\alpha)\KL(p_{k+1}\Vert p_k)\geq0,
    \]
    since $\eta^{-1}\geq L_n+\alpha$.
\end{proof}

\begin{lemma}[Geometric contraction and objective convergence]
    \label{lem:app-geometric-convergence}
    Let $p^\star$ be the unique maximizer established in
    Appendix~\ref{app:regularization}, and write
    $D_k:=\KL(p^\star\Vert p_k)$ and
    $\Delta_k:=G_{n,\alpha}(p^\star)-G_{n,\alpha}(p_k)$.
    Then
    \[
        D_{k+1}\leq\rho D_k,
        \qquad
        \Delta_{k+1}\leq\eta^{-1}(\rho D_k-D_{k+1}),
        \qquad k\geq0.
    \]
    Consequently, $D_k\leq\rho^kD_0$ and
    $\Delta_k\leq\eta^{-1}\rho^kD_0$ for every $k\geq1$.
\end{lemma}
\begin{proof}
    The initial divergence $D_0$ is finite because $p^\star\in\mathcal D$.
    Every subsequent $D_k$ is finite as well, since
    \[
        D_k=\KL(p^\star\Vert p_0)
        -\mathbb E_{p^\star}\!
        \left[\log\frac{p_k(X)}{p_0(X)}\right]
    \]
    and the logarithm is bounded by
    Lemma~\ref{lem:app-well-defined-updates}.
    Taking $q=p^\star$ in \cref{eq:app-linearized-master-bound},
    dropping its final nonpositive term, and using
    $\eta^{-1}-\alpha=\eta^{-1}\rho$ yields
    \[
        0\leq\Delta_{k+1}
        \leq\eta^{-1}(\rho D_k-D_{k+1}).
    \]
    The left inequality follows from maximality of $p^\star$.
    It forces $D_{k+1}\leq\rho D_k$, and iteration gives
    $D_k\leq\rho^kD_0$.
    For $k\geq1$, nonnegativity of $D_k$ also gives
    \[
        \Delta_k\leq\eta^{-1}(\rho D_{k-1}-D_k)
        \leq\eta^{-1}\rho D_{k-1}
        \leq\eta^{-1}\rho^kD_0.
    \]
    If $\rho=0$, the same inequalities imply $D_1=0$ and
    $\Delta_1=0$, so $p_1=p^\star$.
    Otherwise, both bounds decay
    geometrically because $0<\rho<1$.
\end{proof}

The three lemmas establish the following convergence guarantee.

\begin{theorem}[Convergence of exact linearized HVFT]
    \label{thm:app-linearized-convergence}
    Under the standing measurability and boundedness assumptions, let
    $n\geq1$, $\alpha>0$, and initialize at the pretrained distribution
    $p_0$.
    Choose a constant step size
    \[
        0<\eta\leq\frac{1}{L_n+\alpha},
        \qquad \rho:=1-\alpha\eta\in[0,1).
    \]
    Then every update in \cref{eq:app-linearized-update} has a unique
    solution, and $G_{n,\alpha}(p_{k+1})\geq G_{n,\alpha}(p_k)$.
    For the unique maximizer $p^\star$ established in
    Appendix~\ref{app:regularization},
    \begin{align}
        \KL(p^\star\Vert p_k)
        & \leq\rho^k\KL(p^\star\Vert p_0),          &  & k\geq1,
        \label{eq:app-linearized-kl-rate}                         \\
        G_{n,\alpha}(p^\star)-G_{n,\alpha}(p_k)
        & \leq\eta^{-1}\rho^k\KL(p^\star\Vert p_0), &  & k\geq1.
        \label{eq:app-linearized-objective-rate}
    \end{align}
    In particular, $p_k\to p^\star$ in total variation.
    When $\rho=0$,
    the first update already equals $p^\star$.
\end{theorem}

\begin{proof}
    Lemma~\ref{lem:app-well-defined-updates} establishes existence and
    uniqueness of each update.
    Lemma~\ref{lem:app-one-step-bound}
    proves monotonic improvement, and
    Lemma~\ref{lem:app-geometric-convergence} gives
    \cref{eq:app-linearized-kl-rate,eq:app-linearized-objective-rate},
    including the case $\rho=0$.
    Finally, Pinsker's inequality gives
    \[
        \|p_k-p^\star\|_{\mathrm{TV}}
        \leq\sqrt{\frac{\KL(p^\star\Vert p_k)}{2}}
        \leq\sqrt{\frac{\rho^k\KL(p^\star\Vert p_0)}{2}},
    \]
    which proves convergence in total variation.
\end{proof}

At the largest admissible step size, $\eta=(L_n+\alpha)^{-1}$,
the contraction factor is $\rho=L_n/(L_n+\alpha)$.
These guarantees apply to exact first variations and exact optimization
over distributions.
Monte Carlo rewards, neural parameterization, and
approximate inner optimization require additional error analysis.

\section{Practical implementation aspects}\label{app:practicality}

\subsection{Dealing with the log-ratio}

A practical implementation must account for the log-density ratio
$\log p(X)-\log p_0(X)$ in \cref{eq:proximal-update},
which involves marginal densities that are often intractable for diffusion
models. We consider two approaches. The first, used in our experiments,
sets $\alpha=0$ while retaining the local KL penalty relative to $p_k$.
This avoids evaluating the density ratio and applies to both continuous
and discrete generators. The local penalty constrains each update,
while the cumulative departure from $p_0$ depends on the
step sizes $\eta_k$ and the stopping point. This approach is empirically
effective, as shown in \Cref{sec:experiments}.

The second approach, applicable to continuous diffusion and flow-matching models, retains $\alpha > 0$ and uses a solver driven by reward gradients, such as Adjoint Matching \citep{domingoenrich2024adjoint}. The adaptive reward has gradient
\[
\nabla_x\left[
    f_{n,p_k}(x)-\alpha\log\frac{p_k(x)}{p_{0}(x)}
\right]
=
\nabla_x f_{n,p_k}(x)
-
\alpha\left[
    \nabla_x\log p_k(x)
    -
    \nabla_x\log p_0(x)
\right],
\]
wherever the required derivatives exist. This approach requires estimates of the output-distribution scores, which are readily available for diffusion and flow-matching models \citep{domingoenrich2024adjoint}. The only remaining obstacle is the first term: a differentiable approximation of the marginal hypervolume reward $f_{n,p_k}$.

Random hypervolume scalarizations provide a convenient representation of this reward \citep{zhang2020random}. For the objective vector $\rvec(x) \in \mathbb{R}^M$ and reference point $\rref$, define
\[
s_\lambda(\rvec(x)-\rref)
=
\left[
    \min_{1\leq m\leq M}
    \frac{r_m(x)-r_m^{\mathrm{ref}}}{\lambda_m}
\right]_+^M,
\qquad
\lambda\sim\operatorname{Uniform}(S_+^{M-1}),
\]
where $[a]_+ = \max\{a,0\}$ and
\[
S_+^{M-1}
=
\left\{
    \lambda\in\mathbb R_{>0}^M:
    \|\lambda\|_2=1
\right\}.
\]

The marginal hypervolume reward in \cref{eq:variation-hvi} can then be written as
\begin{align}
    f_{n,p_k}(x)
&=
n\,c_M\,
\mathbb E_{\lambda,\,A\sim p_k^{\otimes(n-1)}}
\left[
    \left[
        s_\lambda(\rvec(x)-\rref)
        -
        \max_{x'\in A}s_\lambda(\rvec(x')-\rref)
    \right]_+
\right], \label{eq:marginal-hv-scalarize}\\
c_M&=\frac{\pi^{M/2}}{2^M\,\Gamma(M/2+1)}. \notag
\end{align}

This representation supports Monte Carlo estimation over directions $\lambda$ and background sets $A$: for each sample, draw a direction $\lambda$ and a set $A$ of $n-1$ points from $p_k$, and evaluate the bracketed expression.

A smooth surrogate of \cref{eq:marginal-hv-scalarize} is obtained by replacing the minimum in $s_\lambda$ and each positive-part operation $[a]_+$ with softmin and softplus respectively. The maximum over the background set term $\max_{x'\in A} s_\lambda(\rvec(x')-\rref)$ need not be smoothed, since it is independent of $x$.

Automatic differentiation of this surrogate then yields an estimate of the reward gradient $\nabla_x f_{n,p_k}(x)$ for use in Adjoint Matching, provided that $\rvec$ is itself differentiable. Combined with the score terms above, this furnishes the full gradient of the entropy-regularized reward required by the solver.
\subsection{Computational Cost}

At first glance, Algorithm~\ref{alg:expected-hv-finetuning} appears to require a $K$-fold increase in training computation. In practice, however, the total training budget can be split across the $K$ fine-tuning rounds, so the overall number of gradient steps remains unchanged. \AlgNameShort's only additional cost relative to running the inner solving algorithm for the same number of epochs therefore comes from generating and scoring the background sets—a small overhead relative to the full fine-tuning pipeline. Table~\ref{tab:execution_time} compares \AlgNameShort's runtime to that of its inner solver counterpart in Section~\ref{sec:experiments}. For the TR2-D2 comparison, statistics and standard errors are calculated on 9 runs of peptide tasks. Each run used one 40GB NVIDIA A-100 GPU. For the DiffusionNFT comparison, statistics are computed over 3 runs of the Geometry task. Each run used a 32GB NVIDIA RTX 5000 Ada Generation GPU.

\begin{table}[!h]
    \vspace{10pt}
  \centering
  \small
  \caption{Average execution time in hours. The first block aggregates results across the three peptide tasks. The second block aggregates runs for the Geometry task. The performance of these runs is reported in Section~\ref{sec:experiments}. \AlgNameShort's execution time remains comparable to its inner solver counterpart.}
  \label{tab:execution_time}
  \begin{tabular}{lc}
    \toprule
    Method & Execution time (hours) \\
    \midrule
    TR2-D2 & $12.384 \pm 1.114$ \\
    \AlgNameShort & $10.734 \pm 0.310$ \\
    \midrule
    DiffusionNFT & $6.399 \pm 0.069$ \\
    \AlgNameShort & $7.204 \pm 0.082$ \\
    \bottomrule
  \end{tabular}
\end{table}

\section{Experimental details} \label{app:experimental-details}
\subsection{Metrics} \label{app:metrics}
For each metric, we sample 1000 valid designs from the generator. Since generated designs may not always be valid, we allow repeated sampling until the 1000 quota is fulfilled or a maximum of 10000 trials is exceeded.

Our main objective-space metric is the mean hypervolume of 100 samples from the generator. We estimate the 100-hypervolume with sub-sampling:  we draw \(P=100000\) subsets of size \(100\) without replacement from the pool and average their hypervolumes. For each index $i \in [1, 1000]$, we average the hypervolume values of the permutations that exclude $\rvec(X_i)$ to form a leave-one-out estimate $\widehat{\mathrm{HV}}_{-i}$.  The final hypervolume is the mean of these leave-one-out estimates, $\widehat{\mathrm{HV}} = \frac{1}{N}\sum_i \widehat{\mathrm{HV}}_{-i}$. 10-hypervolume is computed the same way.

We used the full set of (up to) 1000 samples for the design-space metrics. Validity is computed as the number of valid samples divided by the number of attempts required to reach this quota, where validity is determined by problem-specific criteria. Following \cite{Santi2026ActiveFE}, we measure generator diversity via the Vendi score of valid designs \citep{Friedman2022TheVS}.

To assess design-space coverage, we adopt a metric similar to the fixed-threshold clustering metric proposed in the same work. Designs are embedded in a metric space and grouped into clusters such that any two designs within a cluster lie within a fixed radius of one another; the resulting number of clusters can be interpreted as the number of hyperspheres needed to cover an equal-sized set of designs from each method, with more clusters indicating broader coverage. Because this metric depends on the choice of radius — which encodes an arbitrary notion of how "similar" two designs must be to count as the same — we avoid committing to a single threshold. Instead, we compute the number of clusters across 100 equally spaced thresholds and integrate numerically, yielding a single coverage score that is independent of any particular similarity cutoff.

\subsection{Implementation Details} \label{app:implementation}

\AlgNameShort introduces four method-specific parameters: the background count
$B$, the set size $n$, the regularization coefficient $\alpha$, and the number of outer loop steps $K$. The background count controls the precision of the Monte Carlo estimator of $f_{n,p_k}$. We found that $B=85$ provides sufficient accuracy in practice and use this value throughout our experiments. We also keep $n=10$ fixed across experiments and examine its effect in Appendix~\ref{app:ablation}. Following the
discussion in Appendix~\ref{app:practicality}, we set the pretrained-model
regularization coefficient $\alpha=0$, while retaining the local KL
penalty between successive iterates. The local trust region coefficient $\eta_k$ is fixed across $k$ so we drop the subscript in the remainder of this section. $K$ is task specific.

\subsubsection{Therapeutic Peptides} \label{app:implementation-pep}
To compute Vendi score and coverage, we use 1024-bit Morgan fingerprints (radius 2) \citep{morganfp} and Tanimoto similarity. The reference point for hypervolume computation is written in Table~\ref{tab:ref-pep} and is shared for all peptide tasks.

\begin{table}[]
\centering
\caption{Reference points used for hypervolume computation across all peptides tasks.}
\label{tab:ref-pep}
\resizebox{\linewidth}{!}{%
\begin{tabular}{lccccc}
\toprule
\textbf{Property} & Binding Affinity & Permeability & Nonfouling & Solubility & Hemolysis \\
\midrule
\textbf{Reference Point} & 4.5 & -10.0 & 0.0 & 0.0 & 0.0 \\
\bottomrule
\end{tabular}%
}
\end{table}

\paragraph{TR2-D2, TR2-D2++.} Following \citet{tang2025tr2d2}, we finetune the base model for 1000 epochs total, evaluating every 100 epochs. Hyperparameters for TR2-D2 and TR2-D2++ are taken directly from \citet{tang2025tr2d2}.

\paragraph{\AlgNameShort.} \AlgNameShort decomposes this same 1000-epoch budget into $K=50$ linear finetuning subproblems, each run for 20 epochs. Its inner solver is the single-objective variant of TR2-D2, and it shares the same hyperparameters as the baselines except for $\eta^{-1}$. Table~\ref{tab:tr2d2-hyperparams} summarizes these settings.

\begin{table}[]
\centering
\caption{Hyperparameters for TR2-D2 baselines and \AlgNameShort inner-loop finetuning.}
\label{tab:tr2d2-hyperparams}
\resizebox{\linewidth}{!}{%
\begin{tabular}{llc}
\toprule
\textbf{Hyperparameter} & \textbf{Name in \cite{tang2025tr2d2}} & \textbf{Value} \\
\midrule
Num. of Replicates for Weighted Denoising Cross-Entropy & $R$                    & 16 \\
Buffer Size                                             & $B$                    & 20 \\
Sequence Length                                         & $L_\mathrm{seq}$       & 200 \\
Num. of Children                                        & $M$                    & 50 \\
Num. of Decoding Steps                                  & $N_\mathrm{steps}$     & 128 \\
Num. of MCTS Iterations                                 & $N_\mathrm{iter}$      & 20 \\
Resample Every $n$ Steps                                & $N_\mathrm{resample}$  & 20 \\
Max Token Length                                        & $L$                    & 200 \\
Learning rate                                                  & --              & $10^{-4}$ \\
$\eta^{-1}$                                                  & $\alpha$               & 0.1 (TR2-D2) or 1 (\AlgNameShort) \\

\bottomrule
\end{tabular}%
}
\end{table}
\paragraph{Peptune.} To ensure a fair comparison, we depart from the evaluation protocol used for Peptune in \cite{tang2025peptune}. Specifically, we set $N_\mathrm{iter}=200$ and $M=50$, so that at most 10000 samples are considered during inference. Every valid candidate encountered is retained in a running buffer until it reaches 1000 samples. Candidates that are unrolled during the tree search but turn out invalid still count toward the validity total.

\subsubsection{Molecular Design} \label{app:implementation-mol}

In all tasks, we fix the number of atoms to 10. We normalize Synthetic Accessibility via $(10 - SA)/9$ and additionally scale the dipole moment by a factor of 20 to avoid magnitude bias across baselines. Invalid candidates receive a reward of $-1$, giving the finetuning algorithms a signal when a design is invalid. The reference points for the Topology and Geometry tasks are given in Table~\ref{tab:ref-mol}.

All finetuning methods rely on DiffusionNFT, either as the main solver or as a component of their algorithm. Its hyperparameters are held fixed across methods and are summarized in Table~\ref{tab:diffusionnft-hyperparams}.

\paragraph{Topology.} For the topology task, we allow 900 finetuning epochs, evaluating every 50. We compute Vendi score and coverage using 1024-bit Morgan fingerprints (radius 2) with Tanimoto similarity. Here, \AlgNameShort uses $K=9$ to get 100 epochs per inner finetuning subproblem.

\paragraph{Geometry.} For the geometry task, computational constraints limit us to 600 epochs, evaluating every 50. We compute Vendi score and coverage using USRCAT fingerprints and the USRCAT similarity score \citep{Schreyer2012}, as implemented in RDKit \citep{greg_landrum_2026_22140358}. Here, \AlgNameShort uses $K=6$ to get 100 epochs per inner finetuning subproblem.

\paragraph{DiffusionBlend.} \citet{Cheng2025DiffusionBI} do not describe an automatic weight-selection procedure. We generate evenly spread reward-weighting tradeoffs via Das--Dennis sampling (for Geometry) or uniform sampling in $[0,1]$ (for Topology).

\begin{table}[]
\centering
\caption{Reference points used for hypervolume computation for the Topology and Geometry tasks.}
\label{tab:ref-mol}

\begin{tabular}{lccccc}
\toprule
\textbf{Property} & QED & SA & Dipole Moment\\
\midrule
\textbf{Reference Point} & -1.0 & -1.0 & -1.0 \\
\bottomrule
\end{tabular}%

\end{table}

\begin{table}
\centering
\caption{Additional hyperparameters for DiffusionNFT finetuning.}
\label{tab:diffusionnft-hyperparams}

\begin{tabular}{llc}
\toprule
\textbf{Hyperparameter} & \textbf{Name in \cite{Zheng2025DiffusionNFTOD}} & \textbf{Value} \\
\midrule
Guidance Strength                & $\beta$                                     & 50 \\
KL Regularization Coefficient    & Not in paper, but in official implementation    & 0.5 \\
Learning Rate                    & --                                          & 1e-5 \\
\bottomrule
\end{tabular}
\end{table}

\section{Ablations} \label{app:ablation}

To assess \AlgNameShort's sensitivity to $n$, we conduct additional experiments on representative therapeutic-peptide and small-molecule design tasks, holding all other parameters fixed as specified in Appendix~\ref{app:implementation}. Table~\ref{tab:n_ablation} shows substantial hypervolume gains over the pretrained model across the tested values of $n$, which span up to two orders of magnitude. These results suggest that \AlgNameShort is robust to the choice of $n$.

On the Topology task, which has two objectives, increasing $n$ does not improve hypervolume: $n=10$ achieves the highest values. This suggests that relatively few samples may suffice to capture the main trade-offs on this task's Pareto front.

Conversely, on the peptide task TfR, increasing $n$ consistently yields better hypervolume values. With five rewards, the Pareto front of this task is likely more complex and not easily covered by a small set of points. The larger variance at $n=150$ has a simple explanation: Since $B=85$, each point's marginal contribution is estimated over 85 sets, each containing 149 other scores. Contributing additional hypervolume coverage against a set of that size is hard, so most of these 85 estimates are zero, and only a handful are non-zero. As $n$ grows, this sparsity makes the average marginal contribution noisier, producing higher variance across runs.

\begin{table}[]
\centering
\caption{Effect of $n$ on objective- and design-space metrics for the TfR and topology tasks, with the pretrained base model as a reference. Values are means $\pm$ standard errors across three runs. Boldface and underlining indicate the highest and second-highest means within each task, respectively.}
\label{tab:n_ablation}
\setlength{\tabcolsep}{4pt}
\renewcommand{\arraystretch}{1.15}
\resizebox{\textwidth}{!}{%
\begin{tabular}{llccccc}
\hline
& & \multicolumn{2}{c}{Objective-space metrics}
& \multicolumn{3}{c}{Design-space metrics} \\
\cline{3-4}\cline{5-7}
Task & Model / $n$ & 100-Hypervolume & 10-Hypervolume
& Diversity & Coverage & Valid fraction \\
\hline
TfR & Base
& $2.93 \pm 0.04$
& $1.69 \pm 0.01$
& $60.35 \pm 0.14$
& $401.50 \pm 0.28$
& $\mathbf{0.321 \pm 0.007}$ \\
& $10$
& $11.57 \pm 0.16$
& $10.15 \pm 0.18$
& $75.89 \pm 9.42$
& $\mathbf{485.80 \pm 17.27}$
& $\underline{0.250 \pm 0.021}$ \\
& $50$
& $\underline{12.231 \pm 0.153}$
& $\underline{10.589 \pm 0.226}$
& $\underline{93.2 \pm 3.0}$
& $437.9 \pm 62.9$
& $0.210 \pm 0.078$ \\
& $150$
& $\mathbf{13.757 \pm 1.721}$
& $\mathbf{11.107 \pm 0.891}$
& $\mathbf{107.3 \pm 8.6}$
& $\underline{454.1 \pm 43.6}$
& $0.235 \pm 0.114$ \\
\hline
Topology & FlowMol
& $3.116 \pm 0.014$
& $2.665 \pm 0.007$
& $\underline{266.1 \pm 4.9}$
& $\underline{507.1 \pm 6.4}$
& $0.705 \pm 0.008$ \\
& $10$
& $\mathbf{3.416 \pm 0.005}$
& $\mathbf{3.247 \pm 0.004}$
& $154.9 \pm 9.0$
& $351.6 \pm 11.1$
& $\mathbf{0.734 \pm 0.007}$ \\
& $100$
& $\underline{3.405 \pm 0.004}$
& $\underline{3.200 \pm 0.006}$
& $254.5 \pm 3.5$
& $473.1 \pm 5.0$
& $0.675 \pm 0.008$ \\
& $1000$
& $3.362 \pm 0.010$
& $3.082 \pm 0.017$
& $\mathbf{310.6 \pm 6.3}$
& $\mathbf{547.4 \pm 6.5}$
& $\underline{0.708 \pm 0.012}$ \\
\hline
\end{tabular}%
}
\end{table}

\end{document}